\documentclass[final]{cvpr}

\usepackage{mmstyles}
\usepackage{comment}
\usepackage{microtype}
\usepackage{subcaption}
\usepackage{multirow}
\usepackage{color}
\usepackage{times}
\usepackage{epsfig}
\usepackage{graphicx}
\usepackage{amsmath}
\usepackage{amssymb}
\usepackage{wrapfig}
\usepackage{pifont}
\usepackage{adjustbox}
\usepackage{algorithm}
\usepackage{algorithmic}

\newcommand{\cmark}{\ding{51}}

\usepackage{bbm}

\usepackage[pagebackref=true,breaklinks=true,colorlinks,bookmarks=false]{hyperref}

\usepackage{ulem}
\newcommand{\bad}[1]{\textcolor[RGB]{124,75,0}{\uwave{#1}}} 

\newcommand{\iid}{\textit{i}.\textit{i}.\textit{d}.}

\newcommand{\bfit}[1]{\textbf{\textit{#1}}}

\usepackage{multicol}

\usepackage{booktabs}

\def\cvprPaperID{0000} 
\def\confYear{CVPR 2012}

\begin{document}

\title{Positional Encoding as Spatial Inductive Bias in GANs}


\author{Rui Xu$^{1}$ \hspace{9pt} Xintao Wang$^{3}$ \hspace{9pt} Kai Chen$^{4}$ \hspace{9pt} Bolei Zhou$^{1}$ \hspace{9pt} Chen Change Loy$^{2}$ \\
	\small{$^{1}$ CUHK-SenseTime Joint Lab, The Chinese University of Hong Kong \hspace{5pt}
      $^{2}$ Nanyang Technological University} \\ 
   \small{$^{3}$ Applied Research Center, Tencent PCG \hspace{5pt}
		$^{4}$ SenseTime Research} \\ 
   {\tt\small \{xr018, bzhou\}@ie.cuhk.edu.hk \hspace{5pt} xintao.wang@outlook.com} \\
   {\tt\small chenkai@sensetime.com \hspace{5pt} ccloy@ntu.edu.sg}
 }

\maketitle

\begin{abstract}
   \vspace{-8pt}

SinGAN shows impressive capability in learning internal patch distribution despite its limited effective receptive field.
We are interested in knowing how such a translation-invariant convolutional generator could capture the global structure with just a spatially \iid~input.
In this work, taking SinGAN and StyleGAN2 as examples, we show that such capability, to a large extent, is brought by the implicit positional encoding when using zero padding in the generators.
Such positional encoding is indispensable for generating images with high fidelity. The same phenomenon is observed in other generative architectures such as DCGAN and PGGAN. 
We further show that zero padding leads to an unbalanced spatial bias with a vague relation between locations.
To offer a better spatial inductive bias, we investigate alternative positional encodings and analyze their effects.
Based on a more flexible positional encoding explicitly, we propose a new multi-scale training strategy and demonstrate its effectiveness in the state-of-the-art unconditional generator StyleGAN2. 
Besides, the explicit spatial inductive bias substantially improve SinGAN for more versatile image manipulation.

\end{abstract}

\vspace{-15pt}

\section{Introduction}
\label{sec:intro}

\begin{figure}[tb]
  \setlength{\abovecaptionskip}{0pt}
  \centering
  \small
  \includegraphics[width=\linewidth]{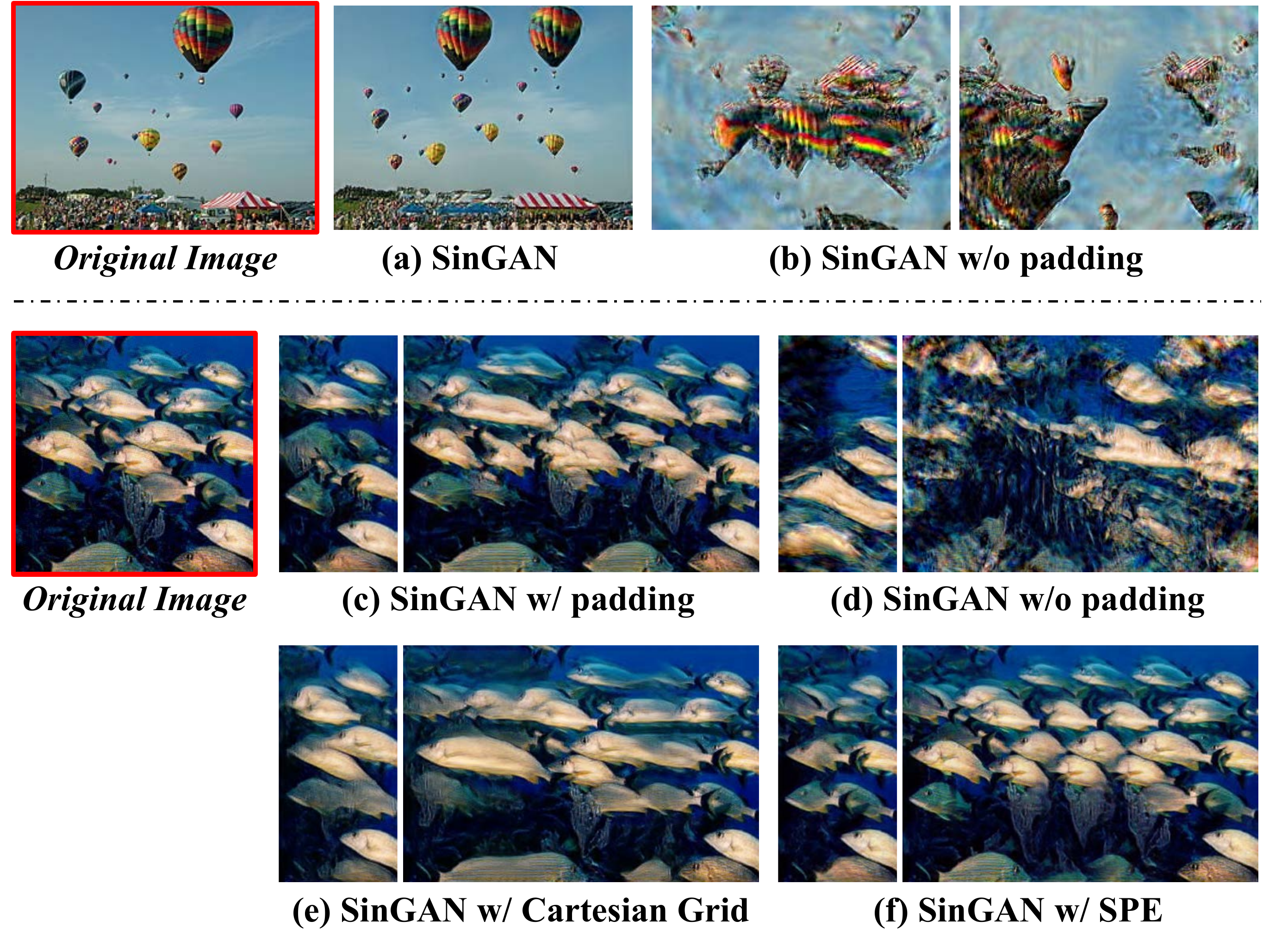}
  \caption{Images sampled from the internal patch distribution learned by SinGAN. Above the dotted line, we present sampled balloons with standard SinGAN and padding-free SinGAN. A more challenging case of generating a school of fish is shown below the dotted line. (c)-(f) show the effects of different positional encodings that we explore on SinGAN.}
  \label{fig:teaser}
  \vspace{-20pt}
\end{figure}

SinGAN~\cite{shaham2019singan} and StyleGAN~\cite{karras2019style,karras2020analyzing} are among the few representative Generative Adversarial Networks (GANs) that show impressive image generative capability.
Both of their generators are based on a fully translation-invariant convolutional network.
One would expect that in an unconditional setting with a spatially \iid~input, the translation invariance property should result in position-agnostic outputs like Fig.~\ref{fig:teaser}(b). Nonetheless, the results of SinGAN shows surprisingly  structured results like Fig.~\ref{fig:teaser}(a).
%

Driven by curiosity, we carefully study this phenomenon and find that it is the zero padding that causes a location-aware bias in the distribution of feature maps. Such a spatial bias gradually spreads from the border to the center of feature maps through the stacked convolutional layers in the generator.
One can regard this spatial bias as an implicit positional encoding, which contributes to the high fidelity of images generated by SinGAN and StyleGAN.
Interestingly, we also observe the same phenomenon in other unconditional generative architectures such as DCGAN~\cite{radford2015unsupervised} and PGGAN~\cite{karras2017progressive}. 

Our observation reveals the importance of introducing positional encoding in generative models.
The original intention of zero padding is to maintain the spatial size of feature maps. It is not specially designed to offer the required spatial inductive bias.
In particular, we find that the bias caused by zero padding is unbalanced over the image space.
Since paddings are introduced at image borders and corners, the positional encoding at those locations is structured. In contrast, the spatial encoding in the center region is highly unstructured due to the gradually diminishing effects of zero padding from borders to the center.
The shortcoming of such bias can be observed from Fig.~\ref{fig:teaser} where we use SinGAN to synthesize an image of a school of fish. In this example, SinGAN generates relatively more structured output at the borders but inferior results at the center of the image.

The aforementioned example suggests the shortcoming of padding in serving the role of positional encoding.
%
The desired positional encoding should keep a consistent spatial structure and be invariant to scale transformation. 
In this study, we investigate two alternatives for explicit positional encoding, \ie, the normalized Cartesian spatial grid~\cite{jaderberg2015spatial} and 2D sinusoidal positional encoding~\cite{devlin2018bert, martin2020nerf, mildenhall2020nerf, vaswani2017attention}, which both guarantee a balanced spatial inductive bias over the whole image space.
We show that these explicit positional encodings allow a convolutional generator to generate images that exhibit a more stable structure and more reasonable patch reoccurrence given an arbitrary scale in the generation, as shown in Fig.~\ref{fig:teaser}(e) and Fig.~\ref{fig:teaser}(f).

With the more flexible explicit positional encodings, we can redesign convolutional generative models to synthesize images at multiple scales even just using a single model.
Achieving this functionality is challenging with existing models. One will typically need to train different generators with different upsampling blocks.
We show that multi-scale generation with a single fully convolutional generator is possible using our newly proposed multi-scale training strategy based on explicit positional encodings. We call it Multi-Scale training with PositIon Encodings (MS-PIE).
%
We demonstrate its effectiveness in the state-of-the-art unconditional generator StyleGAN2.
With MS-PIE, a single StyleGAN2 that is designed for $256\times256$ image generation yields compelling generation quality at multiple scales up to $512\times 512$ or even $1024\times 1024$ pixels, despite that it only contains limited upsampling blocks in its architecture. 
%

We summarize the contributions of this study as follows: (1) we reveal the phenomenon where zero padding unintentionally introduces implicit (but useful) positional encoding in existing convolutional generators. We study this phenomenon through detailed theoretical and empirical analyses. While the influence of padding to translation-invariant convolution has been discussed in recent works~\cite{alsallakh2020mind, islam2020much, kayhan2020translation}, these studies focus on image classification and detection. Our research is the first study that investigates the impacts of such spatial bias on image generation.
(2) We further investigate and present two explicit positional encodings as two new spatial inductive bias in generators, which can substantially improve the versatility and robustness of SinGAN.
(3) We propose a new multi-scale training strategy to achieve high-quality multi-scale synthesis with a single StyleGAN2 that is originally designed for $256\times256$ generation. 
\section{Related Work}



%
\noindent \textbf{Padding Effects.}
Some recent studies~\cite{alsallakh2020mind, islam2020much, kayhan2020translation} discover an intriguing phenomenon in which the widely used zero padding would offer spatial information (an unintended design) in convolutional networks for image classification~\cite{deng2009imagenet,he2016deep} and detection~\cite{he2017mask, ren2015faster}. In general, a spatial bias is induced causing activations at certain locations is systematically elevated or weakened.
%
%
Islam~\etal~\cite{islam2020much} find padding implicitly injects positional information in ResNet~\cite{he2016deep} and VGG~\cite{simonyan2014very}, verified with an auxiliary positional encoding module. 
The experiments in \cite{kayhan2020translation} further show that the effects of padding vary among different architectures~\cite{brendel2019approximating, huang2017densely}.
%
%
%
Alsallakh~\etal~\cite{alsallakh2020mind} observe the similar phenomenon and find that such spatial bias would cause blind spots for detectors~\cite{liu2016ssd}, detrimental to small object detection.
In this work, we show both theoretically and empirically how zero padding accidentally encodes spatial information for convolutional generators.
We find that spatial bias, unlike high-level visual tasks, is actually necessary for generators to work well. We further discuss better choices of spatial inductive bias.
%
The analysis of our study can be easily applied to explain the effects of other padding modes in convolutional generators.

\noindent \textbf{Sinusoidal Positional Encoding.}
Sinusoidal positional encoding (SPE) is widely used in natural language processing (NLP)~\cite{brown2020language, devlin2018bert, vaswani2017attention} and 3D vision~\cite{martin2020nerf,mildenhall2020nerf, tancik2020fourier}.
In the transformer architecture~\cite{vaswani2017attention}, the sequence model relies on SPE to indicate the time step for each word embedding.
SPE provides a stable and reasonable positional encoding for dealing with natural language because the transformation between different time steps in SPE is irrelevant to the length of the input sentence.
%
%
To avoid the spectral bias~\cite{rahaman2019spectral} in the fully connected networks, Martin~\etal~\cite{martin2020nerf} transfers the input features from the low-frequency domain to the high-frequency domain~\cite{tancik2020fourier} through the sinusoidal function.
%
Different from the aforementioned studies, we focus on adopting sinusoidal positional encoding in 2D convolutional generators to obtain more effective spatial inductive bias. 
%

\noindent \textbf{Cartesian Grid.}
Cartesian grid has been introduced in spatial transformer networks~\cite{jaderberg2015spatial} as a standard coordinate system for 2D spatial feature space.
It is widely adopted for differentiable image warping~\cite{Recasens_2018_ECCV, shocher2019ingan, xiao2018spatially} and aligning pixels among different spaces~\cite{dosovitskiy2015flownet, ilg2017flownet, lee2019sfnet, rocco2017convolutional,rocco2018end, zhou2016view}.
In this study, we develop a new role of explicit positional encoding for the normalized Cartesian grid.

\section{Methodology}
\label{sec:method}

As shown in Fig.~\ref{fig:teaser}, once we remove the padding in SinGAN, we observe that the translation-invariant convolutional generator collapses to position-agnostic distribution. 
This suggests that SinGAN relies on zero padding to capture spatial information.
%
%
%
%
%
%
%
From the view of the stochastic process, we clarify how zero padding works as implicit positional encoding.
%
%
After analyzing such implicit positional encoding, we investigate the potential of two explicit positional encodings as better spatial inductive bias in GAN's generator.
%
%
Finally, we present applications on MS-PIE and SinGAN to demonstrate the significance of spatial inductive bias to the generative models.

\subsection{Translation Invariance in Generative Models}
\label{sec:translation-invariance}
%
%
To better understand the effects of zero padding, we first analyze the behavior of padding-free convolutional generators.
%
%
The popular SinGAN adopts a fully convolutional generator with a spatially \iid~noise map as input.
%
%
%
%
Thus, the translation-invariant convolutional network can be regarded as a stochastic process on spatial random variables.
We mainly study two basic statistical properties of the expectation ($\mathbb{E}$) and autocorrelation function ($R$) for the convolutional feature maps.
$\mathbb{E}(y_{\vec{\mathbf{i}}})$ defines the distributional property of each location $\vec{\mathbf{i}}$ in the convolutional feature map $\lbrace y_{\vec{\mathbf{i}}} \rbrace$, while $R(y_{\vec{\mathbf{i}}}, y_{\vec{\mathbf{j}}})$ depicts the relationship between two spatial locations.
Due to the space limitation, a detailed derivation is shown in Appx.~\ref{appx:translation-invariance}.

Taking $x_k \in X_{\vec{\mathbf{i}}}\stackrel{i.i.d.}{\sim} \mathcal{N}(0,1)$ as input, the expectation of the feature map after the first convolutional layer $y_{\vec{\mathbf{i}}}^{(1)}$ is:
{   \setlength{\abovedisplayskip}{10pt}
    \setlength{\belowdisplayskip}{10pt}
    \begin{align}
        \label{eq:e-1}
        \mathbb{E}(y_{\vec{\mathbf{i}}}^{(1)}) & = \sum_{k} w_k^{(1)}\int_{-\infty}^{+\infty}x_kp(x_k)dx_k + b^{(1)} \notag \\
                                & = \sum_k w_k^{(1)} \mathbb{E}(x_k) + b^{(1)} = b^{(1)},
    \end{align}
}
%
\hspace{-0.15cm}where $(w_k^{(1)}, b^{(1)})$ are parameters in the first convolutional layer and the subscript $k$ indicates the $k$-th item in the sum of a convolutional weight ($w_k$) multiplying an input feature ($x_k$).
Besides, $p(\cdot)$ represents the probability density function.
%
As we assume $X_{\vec{\mathbf{i}}}\stackrel{i.i.d.}{\sim} \mathcal{N}(0,1)$, the zero $\mathbb{E}(x_k)$ induces that the expectation of the first convolutional feature is only related to the bias parameter.
After applying a commonly used LeakyReLU function (g) with negative slope ($\gamma$)~\cite{maas2013rectifier} and the second convolutional layer, a more general formulation for the expectation $\mathbb{E}(y_{\vec{\mathbf{i}}}^{(2)})$ should be:
%
%
{   \setlength{\abovedisplayskip}{10pt}
    \setlength{\belowdisplayskip}{10pt}
    \begin{align}
        \label{eq:e-nopad-simple}
        \mathbb{E}(y_{\vec{\mathbf{i}}}^{(2)}) &= \sum_{k} w_k^{(2)}\int_{-\infty}^{+\infty} g(y^{(1)}_k) p(y^{(1)}_k)dy^{(1)}_k + b^{(2)} \notag \\
                &= \sum_k w_k^{(2)} \cdot (\gamma\mathbb{C}_1 + \mathbb{C}_2) + b^{(2)},
    \end{align}
}
%
%
\hspace{-0.15cm}where $\mathbb{C}_1, \mathbb{C}_2$ are constants from the finite piecewise-defined integration.
Eq.~\eqref{eq:e-nopad-simple} shows that the convolutional features keep a spatially identical expectation value, which is a linear combination of the convolutional weights.
%
%
Furthermore, the analysis in the autocorrelation function\footnote{We discard the bias term since the identical addition term does not influence the final conclusion. See complete derivations in the appendix.} further shows that the relation of two positions in the first convolutional feature map is decoupled with the absolute position:  
{\setlength{\abovedisplayskip}{6pt}
\setlength{\belowdisplayskip}{6pt}
\begin{align}
    R(y_{\vec{\mathbf{i}}}, y_{\vec{\mathbf{j}}}) &= \mathbb{E}(y_{\vec{\mathbf{i}}} y_{\vec{\mathbf{j}}}) \notag \\
                \label{eq:r-nopad-simple}
                &= \sum_{x_l \in X_{\vec{\mathbf{i}}} \cap X_{\vec{\mathbf{j}}}} w_{k_l}w_{t_l} \mathbb{E}(x_l^2) \\
                &= R(\vec{\mathbf{i}} - \vec{\mathbf{j}}). \notag
\end{align}
}
%
\hspace{-0.15cm}Here, the condition $x_l \in X_{\vec{\mathbf{i}}} \cap X_{\vec{\mathbf{j}}}$ determines whether $x_l$ belongs to the intersection region of related input features. 
If there is no intersection between two input features, the sum in Eq.~\eqref{eq:r-nopad-simple} will be zero.
Importantly, the intersection region $X_{\vec{\mathbf{i}}} \cap X_{\vec{\mathbf{j}}}$ is determined by the offset vertor $\vec{\mathbf{i}} - \vec{\mathbf{j}}$. Thus, the autocorrelation function of $(y_{\vec{\mathbf{i}}}, y_{\vec{\mathbf{j}}})$ is only related to $\vec{\mathbf{i}} - \vec{\mathbf{j}}$ but irrelevant to the absolute position vector $\{ \vec{\mathbf{i}}, \vec{\mathbf{j}} \}$. 
This proves that after convolution, the features can be regarded as a spatial \bfit{weak stationary stochastic process}.


An essential property of weak stationarity is that the absolute positional information is lost. 
As shown in Fig.~\ref{fig:teaser}(b), without any spatial bias, convolutional generators fail to capture faithful spatial structures, \eg, the position of the ground and the spatial organization of related patches (balloons).
However, it can still output some reasonable patches like balloon texture patterns.
The reason is that the truncated $R(y_{\vec{\mathbf{i}}}, y_{\vec{\mathbf{j}}})$ models the relationship between convolutional features within the limited effective receptive field.
In conclusion, the translation invariance in convolution leads to weak stationarity in features.

\subsection{Padding as Spatial Inductive Bias}
\label{sec:mechanism-pad}
%
As discussed in Sec.~\ref{sec:translation-invariance}, the translation-invariant convolution causes positional information loss from the convolutional features.
Thus, SinGAN should have generated results without reasonable spatial structures like Fig.~\ref{fig:teaser}(b). 
However, as shown in Fig.~\ref{fig:teaser}(a), zero padding unintentionally enables SinGAN to capture the spatial structure of the sky, ground, and \etc.
%
Based on the analysis in Sec.~\ref{sec:translation-invariance}, we will clarify this phenomenon theoretically.
%

\begin{figure}[tb]
    \setlength{\abovecaptionskip}{3pt}
    \centering
    \includegraphics[width=\linewidth]{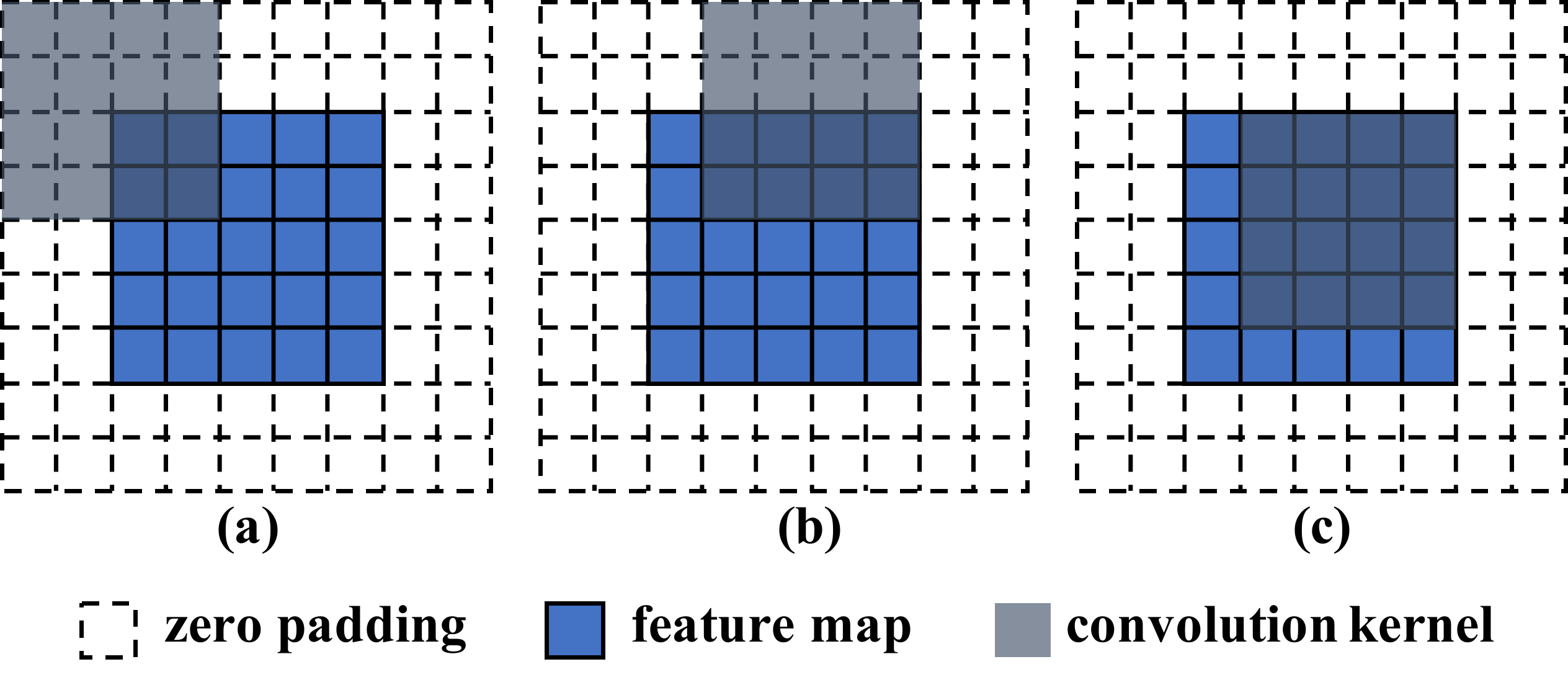}
    \caption{Illustration for the convolutional procedure with zero padding. We move the padding in each layer to the input feature and regard the whole convolutional network as a convolutional layer with a large kernel.}
    \label{fig:conv-pad}
    \vspace{-10pt}
\end{figure}

%
From the view of the effective receptive field~\cite{luo2016understanding}, we regard the whole convolutional network as a convolutional layer with a large kernel and move all of paddings to the input, which is illustrated in Fig.~\ref{fig:conv-pad}. 
Then, the linear combination of the convolutional kernel weights in Eq.~\eqref{eq:e-nopad-simple} and Eq.~\eqref{eq:r-nopad-simple} will be influenced by zero padding:
{\setlength{\abovedisplayskip}{8pt}
\setlength{\belowdisplayskip}{8pt}
\begin{align}
    \label{eq:e-pad}
    \mathbb{E}(y_{\vec{\mathbf{i}}}) &= \sum_k w_k (\gamma\mathbb{C}_1 + \mathbb{C}_2) \mathbbm{1}(x_k\notin Pad) + b, \\
    \label{eq:r-pad}
    R(y_{\vec{\mathbf{i}}}, y_{\vec{\mathbf{j}}}) 
                &=  \sum_{x_l \in X_{\vec{\mathbf{i}}} \cap X_{\vec{\mathbf{j}}}} w_{k_l}w_{t_l} \mathbb{E}(x_l^2) \mathbbm{1}(x_l \notin Pad),
\end{align}
}
\hspace{-0.1cm}where the indicator function $\mathbbm{1}(x_i \notin Pad)$ determines if the current input belongs to the padding regions.
%
%
Intuitively, when the convolution kernel meets input features containing zero padding, some convolutional weights will be multiplied by the zero value.
%
The number of such inevitable zero terms varies as the convolution kernel slides over the feature map. 
Namely, the padding effects on Eq.~\eqref{eq:e-pad} and Eq.~\eqref{eq:r-pad} are determined by the overlap of the convolution kernel and zero padding.
%
%
Therefore, zero padding implicitly injects positional information through the location-variant $\mathbb{E}(y_{\vec{\mathbf{i}}})$ and $R(y_{\vec{\mathbf{i}}}, y_{\vec{\mathbf{j}}})$.
%
The $\mathbb{E}(y_{\vec{\mathbf{i}}})$ in Eq.~\eqref{eq:e-pad} yields the position-aware distribution of the spatial random variables, which is a kind of implicit positional encoding.
%
%
In Appx.~\ref{sec:appx-leak}, we will show that the location-variant $R(y_{\vec{\mathbf{i}}}, y_{\vec{\mathbf{j}}})$ can be applied to explain the behavior of other padding modes.
%
%

%
%
%
%
%

\subsection{Analysis on Implicit Positional Encoding}
\label{sec:analysis-pad}
%
The implicit positional encoding introduced by zero padding offers unbalanced spatial inductive bias over the whole image space.  
%
As shown in Eq.~\eqref{eq:e-pad}, the top-left corner in Fig.~\ref{fig:conv-pad} will receive a unique position encoding, because such an overlap of zero padding and the convolutional kernel must only exist at the top-left corner.
However, as the convolution kernel slides away from corners and borders, the central locations (Fig.~\ref{fig:conv-pad}(c)) will be encoded with the same positional information.
%
%
%
An important property of this implicit positional encoding is that distinct \bfit{spatial anchors} provide fixed yet definite spatial bias at corners.
Nevertheless, the distance between two positional encodings becomes uncertain (or even zero) in the central regions.
Following the definition in transformer~\cite{vaswani2017attention}, we call such distance as \bfit{transformation between locations}, because we mainly care about how to transform from one location to another one in the positional encoding.
%

\begin{figure}[tb]
    \centering
    \includegraphics[width=\linewidth]{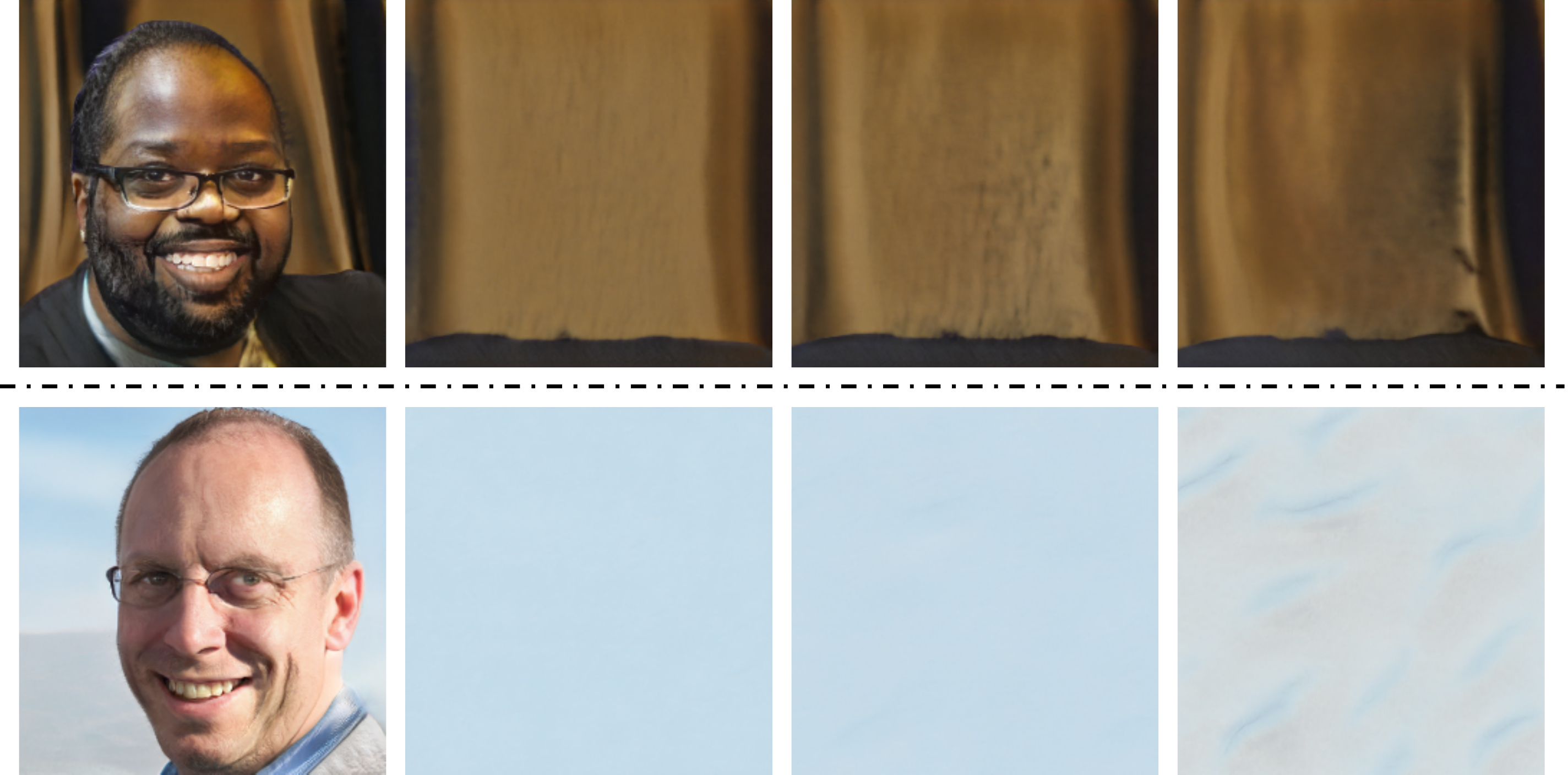}
    \caption{Images sampled from standard StyleGAN2 (above the dotted line) and padding-free StyleGAN2 (under the dotted line). The first column is sampled with the original learned constant input. The other three columns are sampled with different identical values (from left to right: 0, 0.5, 1) filling in the learned constant input at the start of the generator.}
    \label{fig:pad-free}
    \vspace{-10pt}
\end{figure}

%
Such unbalanced spatial inductive bias is not ideal for image generation.
%
Intuitively, without a clear transformation between locations, the uniqueness of the positional encoding cannot be guaranteed.
%
%
Consequently, the convolutional generator fails to precisely portray the desired objects at the center of the generated image, as shown in Fig.~\ref{fig:teaser}(c).
Furthermore, implicit spatial anchors also bring frozen structures at borders.
To show this padding effect on StyleGAN2, we fill in the constant input ahead of its convolutional generator with an identical value.
The identical constant input should have caused spatially consistent patterns because of the weak stationarity in convolutional features. 
Nevertheless, Fig.~\ref{fig:pad-free} shows generated images with borders of similar structure.
Such frozen structures suggest the strong influence from zero padding, which causes the generator to overfit several spatial structures in the training distribution.  
Once the padding is removed, spatially identical color or pattern will cover the whole image, as shown in the second row of Fig~\ref{fig:pad-free}.
%
%
In addition, a shift from precision to recall~\cite{karras2020analyzing, kynkaanniemi2019improved} is observed in our experiments on the padding-free StyleGAN2, suggesting that zero padding actually limits the diversity of a generative model.


%
Prior to SinGAN and StyleGAN, generators typically use a fully connected layer~\cite{goodfellow2014generative} to take the noise vector as input.\footnote{Although, in some implementations, they use transposed convolution on $1\times 1$ noise vector, it equals to applying a linear layer mathematically.} 
Followed by a reshaping layer, this operation explicitly injects spatial information to the feature map ahead of convolutional blocks~\cite{arjovsky2017wasserstein,brock2018large}.
Our experiments show that DCGAN~\cite{radford2015unsupervised} and PGGAN~\cite{karras2017progressive} still rely on the implicit positional encoding to a much greater extent than we expect. 
%
%
%

\subsection{Explicit Positional Encoding for GANs}
\label{sec:pos-emb}

The analysis in Sec.~\ref{sec:analysis-pad} clarifies that implicit positional encoding cannot provide a balanced spatial inductive bias and cannot keep spatially consistent transformation between positions (Fig.~\ref{fig:multi-pe}(a)).
In this section, we will discuss three explicit positional encodings and analyze the spatial inductive bias introduced by them.

\noindent \textbf{Learnable Constant Input.}
In StyleGAN~\cite{karras2019style,karras2020analyzing}, they adopt a $4\times 4 \times 512$ learnable constant as the input of the convolutional generator.
The learnable constant input is fixed across samples but it offers a unique $512$-dimension vector as a learned positional encoding for the $4 \times 4$ input space.
%
However, the chaotic structures in Fig.~\ref{fig:multi-pe}(b) illustrate that the spatial inductive bias defined by the learnable constant input is unclear and lacks explicit priors on image space.

\noindent \textbf{Cartesian Spatial Grid.}
%
%
The Cartesian spatial grid (CSG), used in spatial transformer network~\cite{jaderberg2015spatial}, can play a role of positional encoding.
To avoid large value at huge input space, the Cartesian spatial grid mentioned in this work is normalized as:
{\setlength{\abovedisplayskip}{6pt}
\setlength{\belowdisplayskip}{6pt}
\begin{equation}
    \label{eq:norm-cartesian}
    \vec{P}_{CSG}(i, j) =  2 \cdot [\frac{i}{H} - \frac{1}{2}, \frac{j}{W} - \frac{1}{2}],
\end{equation}
}
\hspace{-0.15cm}where $(i, j)$ represents the spatial location in the $H\times W$ input space.
%
Thus, the corners and central points are fixed to a constant vector, \eg, $[-1, -1]$ for the top-left corner and $[0, 0]$ for the central point.
Such fixed and distinct reference points provide spatial anchors across the image space.
Besides, the transformation between locations is spatially consistent at a single scale:
{\setlength{\abovedisplayskip}{6pt}
\setlength{\belowdisplayskip}{6pt}
\begin{equation}
    \label{eq:trans-csg}
    \vec{P}_{CSG}(i, j) =  2 \cdot [\frac{i - i'}{H}, \frac{j-j'}{W}] + \vec{P}_{CSG}(i', j').
\end{equation}
}
\hspace{-0.1cm}However, the transformation in Eq.~\eqref{eq:trans-csg} is related to the input scale $(H, W)$.
As shown in Fig.~\ref{fig:multi-pe}(c), a larger input scale will bring closer distances between adjacent positional encodings, despite that spatial anchors are fixed.
%
%
Thus, the Cartesian grid is robust to align global structures across multiple scales (Fig.~\ref{fig:teaser}(e)), but the detailed structure will be interpolated similarly with the resizing mode of the Cartesian grid. 

\begin{table}[tb]
    \setlength{\abovecaptionskip}{3pt}
    \centering
    \caption{Summary of spatial inductive bias defined by different positional encodings. `Identical Transform' with `SS' means the transformation between locations is spatially identical at a single scale. The `MS' row shows whether the transformation is scale-invariant. `Interp' is the traditional interpolation while `Expand' denotes the expansion in SPE.}
    \begin{adjustbox}{width=\columnwidth,center}
    \begin{tabular}{c|c|c|c|c|c}
    \bottomrule 
    \multicolumn{2}{c|}{\textbf{Spatial Inductive Bias}} & Pad & \begin{tabular}[c]{@{}c@{}}Learnable \\ Constant \end{tabular} & \begin{tabular}[c]{@{}c@{}}Cartesian\\ Grid\end{tabular} & SPE \\ \hline \hline
    \multicolumn{2}{c|}{Spatial Anchors} &  \cmark &  & \cmark &  \\ \hline
    \multirow{2}{*}{\begin{tabular}[c]{@{}c@{}}Identical\\ Transform\end{tabular}} & SS &  &   & \cmark & \cmark \\ \cline{2-6} 
                                                                                    & MS &  &   &  & \cmark    \\ \hline
    \multirow{2}{*}{\begin{tabular}[c]{@{}c@{}}Resize\\ Mode\end{tabular}}          & Interp &  & \cmark & \cmark & \cmark \\ \cline{2-6} 
                                                                                    & Expand &  &  &  &  \cmark \\ \toprule 
    \end{tabular}
    \end{adjustbox}
    \vspace{-10pt}
    \label{tab:prop-embed}
\end{table}

\begin{figure}[tb]
    \setlength{\abovecaptionskip}{3pt}
    \centering
    \small
    \includegraphics[width=\linewidth]{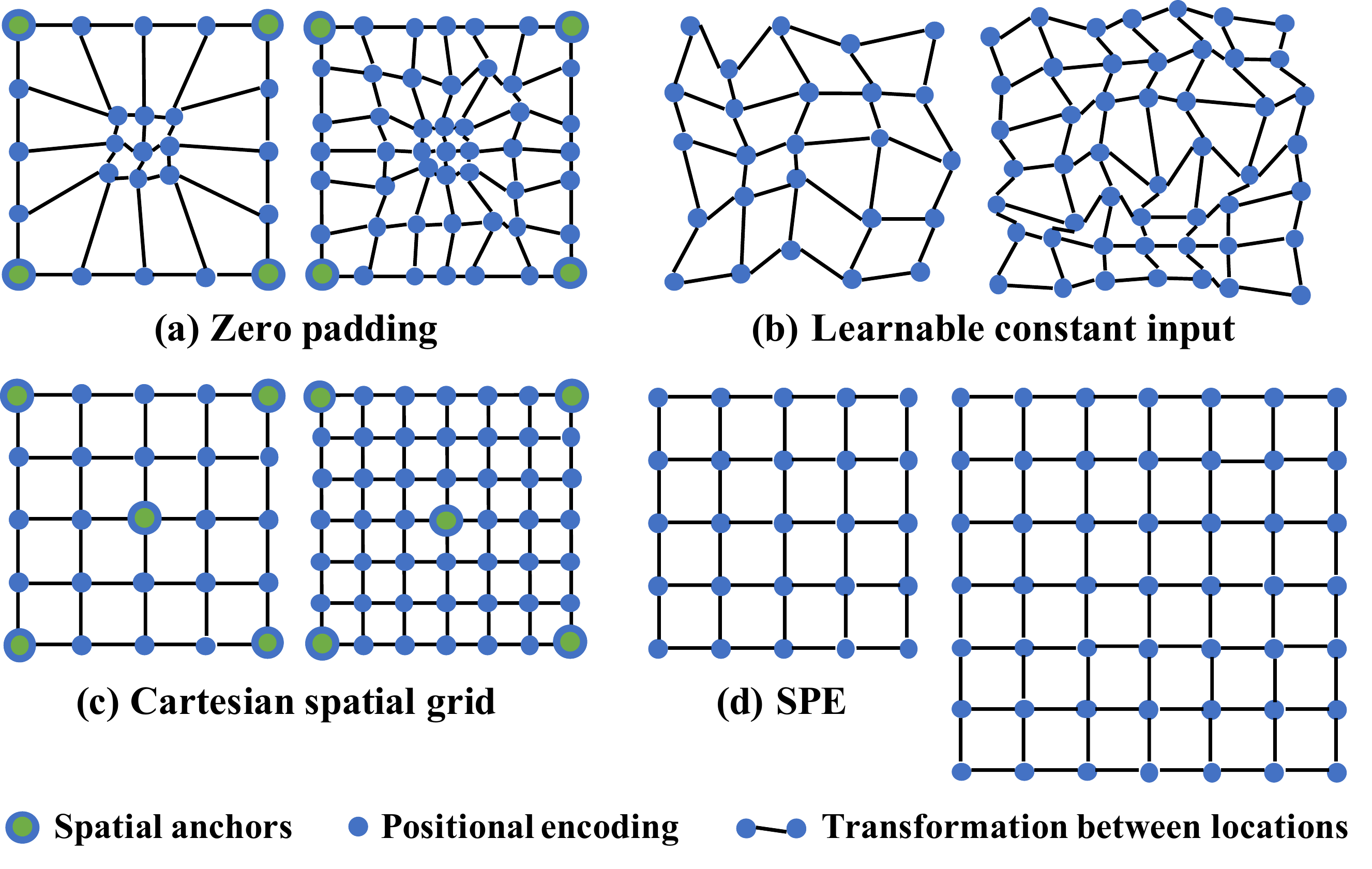}
    \caption{Illustration for the 2D spatial space defined by different positional encodings. For each encoding, we present two spatial spaces at $5\times 5$ and $7\times 7$ scale. (d) shows how SPE naturally expands its space with consistent transformation.}
    \label{fig:multi-pe}
    \vspace{-10pt}
\end{figure}

\noindent \textbf{Sinusoidal Positional Encoding.}
%
Sinusoidal Positional Encoding (SPE) has been widely adopted in NLP~\cite{devlin2018bert, vaswani2017attention} and 3D vision~\cite{martin2020nerf, mildenhall2020nerf}.
To construct 2D positional encoding, we concatenate the encodings in height and width dimensions:
{\setlength{\abovedisplayskip}{6pt}
\setlength{\belowdisplayskip}{6pt}
\begin{equation}
    \label{eq:spe}
    \begin{aligned}
    &[ \underbrace{sin(\omega_0 i), cos(\omega_0 i), \cdots}_{height\; dimension},\; \underbrace{sin(\omega_0 j), cos(\omega_0 j), \cdots}_{width\; dimension}  ], 
    \end{aligned}
\end{equation}
}
\hspace{-0.1cm}where $\omega_k=1/10000^{2k/d}$ and $d$ denotes half of the total encoding dimension. The formulation in Eq.~\eqref{eq:spe} guarantees that the transformation between locations is decoupled with input scales and only related to the offset positional vector:
{\setlength{\abovedisplayskip}{4pt}
\setlength{\belowdisplayskip}{4pt}
\begin{equation}
    \scriptstyle
    \label{eq:trans-spe}
    \begin{bmatrix}
        sin(\omega_k i) \\ cos(\omega_k i)
    \end{bmatrix}
    =
    \begin{bmatrix}
        cos(\omega_k \phi) & sin(\omega_k\phi) \\
        -sin(\omega_k\phi) & cos(\omega_k\phi)
    \end{bmatrix}
    \cdot
    \begin{bmatrix}
        sin(\omega_k i') \\ cos(\omega_k i')
    \end{bmatrix},
\end{equation}
}
\hspace{-0.1cm}where $\phi=i-i'$ is the positional offset.
Thus, without spatial anchors, SPE can naturally expand its space by extending more pixels while keeping the consistent transformation between adjacent positions, as illustrated in Fig.~\ref{fig:multi-pe}(d). 
With such a scale-agnostic transformation, the detailed structure will not be affected when we change the input scale.
Thanks to the stable spatial inductive bias, SPE shows impressive capability in constructing realistic patch recurrence (Fig.~\ref{fig:teaser}(f)).

Table \ref{tab:prop-embed} summarizes the spatial inductive bias that is contained in different positional encodings.  
%
In the following two sections, we will present two applications with various positional encodings and further discuss the significance of spatial inductive bias to convolutional generators.

\subsection{Multi-scale Training with Positional Encoding}
\label{sec:msstylegan}


%
As shown in Tab.~\ref{tab:prop-embed}, an explicit positional encoding can be resized to different scales by either interpolation or expansion.
%
%
Inspired by this property, we derive a new training strategy for performing multi-scale synthesis with a single fully convolutional generator.
Typically, one usually fixes the input scale, like $4\times 4$, and depends on the different number of upsampling blocks to generate multi-scale images.
%
%

Contrary to the above practice, we show that by resizing the explicit positional encoding ahead of the convolutional generator, one can generate images with compelling quality at multiple scales. We call our method as Multi-Scale training with Positional Encoding (MS-PIE).
%
Based on $256^2$ StyleGAN2, we demonstrate the effectiveness of MS-PIE and present the impacts of spatial inductive bias in the padding-based and padding-free settings.
%
%
%
Directly adopting the original learnable constant input in StyleGAN2 causes inferior generation quality due to the lack of any explicit priors on the dynamic image space.
%
%
%
With the standard StyleGAN2 containing zero padding, the explicit scale-invariant transformation (Eq.~\eqref{eq:trans-spe}) in SPE provides a much precise spatial inductive bias over the dynamic input space. 
Therefore, the generator designed for the scale of $256^2$ succeeds in high-quality image synthesis at multiple scales up to $512\times 512$ or even $1024\times 1024$ pixels.
As for the padding-free setting, we discover that the fixed spatial anchors in CSG are essential for the generator to align the global structures among different scales. 
It is the explicit spatial anchors that mitigate the effects of removing zero padding in each layer, which leads to superior performance in the padding-free setting. 

The ultimate goal of MS-PIE is to effectively leverage different image scales for high-fidelity image generation.
%
%
Intuitively, the spatial structure can be efficiently captured in low-resolution space.
The spatial inductive bias guides the generators to enlarge image space according to the resizing mode of the input positional encoding.
Meanwhile, thanks to the priors on dynamic image space and the mixed training of multi-scale images, the high-resolution texture space can be transferred from the low-resolution domain efficiently.
Thus, in our MS-PIE, the generative model is trained on high-resolution images with fewer iterations. 
In each iteration, we sample the current training scale according to a given probability where the higher probability is set for the lower scale.
Then, the real image resolution and the size of the explicit input positional encoding are modified accordingly. 
Besides, to keep the input dimension of the last linear layer in the discriminator unchanged, we insert a $2\times 2$ adaptive average pooling layer~\cite{he2015spatial} before the last linear layer.
%
%
%
%
%


\subsection{SinGAN with Positional Encoding}
\label{sec:singan-pe}
%
%
As discussed in Sec.~\ref{sec:mechanism-pad}, spatial information leaked by zero padding enables SinGAN to capture the global structure and organize various texture patches. 
%
However, the implicit positional encoding defined by zero padding introduces unbalanced spatial inductive bias over the whole image space, which always causes unstructured results in the central region of images (Fig.~\ref{fig:teaser}(a)).
%
This makes it less ideal for applications that require multi-scale internal sampling, \eg, stretching objects with the main structures retained or with reasonable texture patch recurrence~\cite{kwatra2003graphcut, michaeli2014blind}.

The spatial anchors in the Cartesian grid can easily keep the global structures fixed in multi-scale sampling, despite that the contents will be interpolated similarly with the resizing mode of the CSG.
On the other hand, the scale-invariant transformation between locations in SPE guarantees the organization of patches, which leads to a realistic patch recurrence as in Fig.~\ref{fig:teaser}(f).  
To fulfill different requirements, we adopt the positional encoding in SinGAN by sampling on a positional aligned noise distribution.
As the default noise distribution is $\mathcal{N}(0, 1)$, this can be implemented by adding the positional encoding with a sampled noise map.
%
%

\vspace{-5pt}
\section{Experiments}

\vspace{-3pt}
\subsection{Implementation Details}

\noindent \textbf{Remove Padding.} 
%
%
For convolutional generators, the general idea for discarding paddings is to adopt an upsampling layer that interpolates a feature map to a larger size covering the extra padding size for the consecutive convolutional layers.
As for the input convolutional block without any upsampling layers, a larger input map will be adopted to avoid additional paddings.
As the implicit zero padding in the transposed convolutional layer cannot be removed, following PGGAN~\cite{karras2017progressive}, we replace it with a bilinear upsampling layer and a convolutional layer.

\noindent \textbf{Architectures and Training Configurations.}
In addition to internal generative model SinGAN~\cite{shaham2019singan},  we also study various unconditional generator architectures, including DCGAN~\cite{radford2015unsupervised}, PGGAN~\cite{karras2017progressive}, and StyleGAN2~\cite{karras2020analyzing}. 
All of these methods are trained on 8 Tesla V100 GPUs in PyTorch~\cite{NEURIPS2019_9015}. 
We follow their training configurations as closely as possible.
In Sec.~\ref{sec:exp-ms}, we verify the effectiveness of our MS-PIE in $256^2$ StyleGAN2.
Three different scales are adopted in MS-PIE with a sampling probability of $[0.5, 0.25, 0.25]$ and the lower resolution is set with the higher probability.
Other implementation details are specified in the following sections and the supplementary material.

%

\subsection{Effects of Padding in Existing GANs}

\begin{table}[tb]
    \setlength{\abovecaptionskip}{3pt}
    \centering
    \small
    \caption{Results based on $256^2$ StyleGAN2 with a channel multiplier of two~\cite{karras2020analyzing} in FFHQ dataset~\cite{karras2019style, kazemi2014one}. The Fr\'{e}chet inception distance (FID) are reported on two best snapshots before the discriminator has been shown with 10M and 20M images, respectively. The precision and recall are reported on the training snapshot with best FID. $\uparrow$ indicates higher is better, and $\downarrow$ indicates lower is better.}
    \begin{adjustbox}{width=\columnwidth,center}
    \begin{tabular}{l|cc|c|c}
    \bottomrule
    \multirow{2}{*}{\textbf{Training configuration}} & \multicolumn{2}{c|}{FID@256$\downarrow$} & \multirow{2}{*}{\begin{tabular}[c]{@{}c@{}}Precision\\ (\%)$\uparrow$\end{tabular}} & \multirow{2}{*}{\begin{tabular}[c]{@{}c@{}}Recall\\ (\%)$\uparrow$\end{tabular}} \\
                                            & 10M           & 20M          &     &         \\ \hline \hline
    (a) StyleGAN2-C2-256                   & 6.32         & \textbf{5.62}         & \textbf{76.85}  & 50.41     \\
    (b) Deconv $\rightarrow$ Up-Conv     & \textbf{5.79}   & 5.68         & 76.78  &  50.46          \\ \hline
    (c) + Remove padding     &  6.45       & 6.13         & 73.71          & 51.73           \\ 
    (d) +  Cartesian grid   &   6,31     &  6.07    & 73.01   &  \textbf{52.90}      \\
    (e) +  SPE              &   6.40       &  5.86         &  72.94    & \textbf{52.87}      \\ \toprule
    \end{tabular}
    \end{adjustbox}
    \label{tab:pad-stylegan2}
    \vspace{-10pt}
\end{table}

\begin{table}[tb]
    \setlength{\abovecaptionskip}{3pt}
    \centering
    \small
    \caption{Multi-level Sliced Wasserstein Distance (SWD)~\cite{karras2017progressive} between the synthesized and training images in the $128\times 128$ cropped CelebA dataset. Each column in SWD represents one level of Laplacian pyramid~\cite{burt1983laplacian} and the last one shows an average of the three distances. $\downarrow$ indicates lower is better.}
    \begin{tabular}{l|cccc}
    \bottomrule
    \multirow{2}{*}{\textbf{Training configuration}}  & \multicolumn{4}{c}{SWD ($\times 10^3) \downarrow$}  \\
                                            & 128           & 64          & 32          & Avg.    \\ \hline \hline
    (a) PGGAN                               & \textbf{3.162}& \textbf{4.285}& \textbf{5.000}& \textbf{4.149} \\
    (b) + Remove padding                    & 11.169        & 6.945       & 7.488       & 8.534     \\
    (c) + SPE, w/o padding                   & 4.555         & 6.164       & 6.365       & 5.694   \\ \toprule
    \end{tabular}
    \label{tab:pggan-res}
    \vspace{-15pt}
\end{table}

\noindent \textbf{StyleGAN2.}
In Tab.~\ref{tab:pad-stylegan2}, by measuring Fr\'{e}chet inception distantance (FID)~\cite{heusel2017gans} and the precision and recall~\cite{kynkaanniemi2019improved}, we investigate how zero padding influences StyleGAN2~\cite{karras2020analyzing} on FFHQ dataset~\cite{karras2019style}. 
%
%
%
We first switch to a new baseline Tab.~\ref{tab:pad-stylegan2}(b) where transposed convolutions are replaced with a bilinear upsampling layer and a convolutional layer.
This modification yields marginal effects on the final results.
Thus, based on it, we conduct the following experiments on padding-free StyleGAN2.
In Tab.~\ref{tab:pad-stylegan2}(c), the learnable constant input provides spatial information to the convolutional generator.
In Tab.~\ref{tab:pad-stylegan2}(d) and (e), we substitute the constant input with the Cartesian spatial grid and sinusoidal positional encoding, respectively.

As shown in Tab.~\ref{tab:pad-stylegan2}, discarding the implicit positional encoding will directly lead to a higher FID, suggesting that StyleGAN2 actually relies on zero padding to obtain spatial information.
On the other hand, the shift from precision to recall indicates that removing padding allows the generator to explore more reasonable spatial structures. 
Thanks to the spatially consistent transformation between locations, the Cartesian grid and SPE perform better in both FID and recall than the learnable constant input (Tab.~\ref{tab:pad-stylegan2}(c)).
%

\begin{table}[tb]
    \setlength{\abovecaptionskip}{3pt}
    \centering
    \small
    \caption{Multi-level SWD~\cite{karras2017progressive} between the synthesized and training images in the $128\times 128$ LSUN Bedroom dataset. $\downarrow$ indicates lower is better.}
    \begin{tabular}{l|cccc}
    \bottomrule
    \multirow{2}{*}{\textbf{Training configuration}} & \multicolumn{4}{c}{SWD ($\times 10^3$)$\downarrow$} \\
                                            & 128            & 64            & 32            & Avg.           \\ \hline \hline
    (a) DCGAN w/ conv                      & \textbf{11.82}  & 17.30& 28.10 & 19.07         \\ 
    (b) + No padding                        & 22.21          & 27.26       &  44.24          &  31.24       \\ 
    (c) + SPE, w/o padding                   & 14.75          & \textbf{16.52} & \textbf{22.53} & \textbf{17.93}          \\ \toprule
    \end{tabular}
    \label{tab:dcgan-lsun}
    \vspace{-10pt}
\end{table}

\begin{figure}[tb]
    \setlength{\abovecaptionskip}{-3pt}
    \centering
    \includegraphics[width=\linewidth]{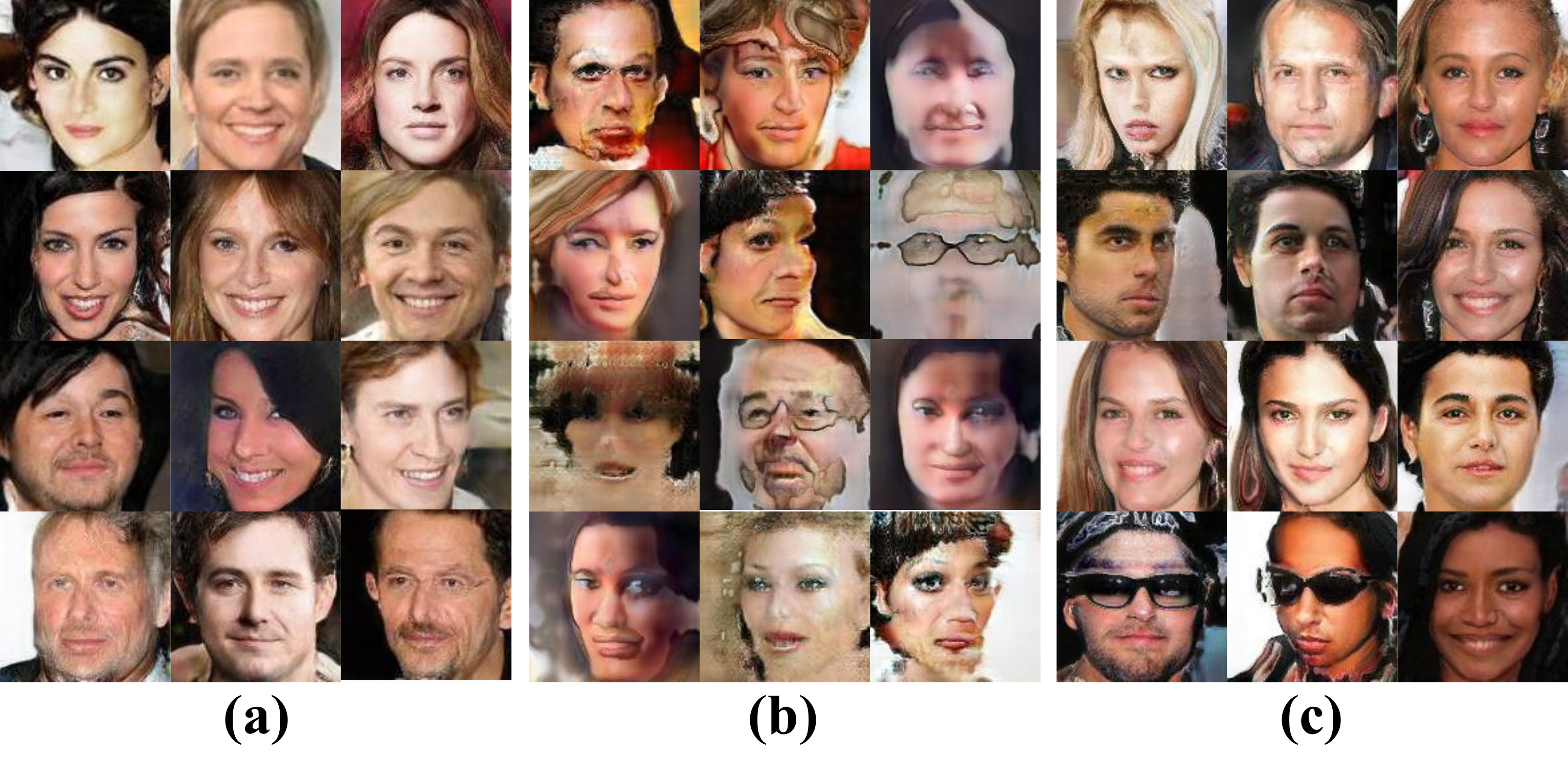}
    \caption{Images sampled from various PGGANs trained on cropped CelebA. (a), (b), and (c) indicate the different training configurations in Tab.~\ref{tab:pggan-res}.}
    \label{fig:exp-pggan-celeba}
    \vspace{-20pt}
\end{figure}

\noindent \textbf{PGGAN and DCGAN.}
We select two popular generator architectures, \ie, PGGAN~\cite{karras2017progressive} and DCGAN~\cite{radford2015unsupervised}, to verify that convolutional generators instinctively obtain implicit positional information from zero padding.
Table \ref{tab:pggan-res} presents the results of the more effective PGGAN on highly-structured cropped CelebA~\cite{liu2015faceattributes} dataset.
In Tab.~\ref{tab:dcgan-lsun}, to remove zero padding, we use the DCGAN architecture with upsampling and convolutional layers as the baseline in the experiments on LSUN-Bedroom dataset~\cite{yu2015lsun}.

Unlike the SinGAN and StyleGAN, PGGAN and DCGAN adopt a linear layer on the input noise vector so that the feature map ahead of the convolutional networks contains spatial information.
However, as shown in Tab.~\ref{tab:pggan-res} and Tab.~\ref{tab:dcgan-lsun}, we still observe a significant increase in SWD at each level after removing zero padding in generators.
%
%
In addition, the padding-free generator fails to synthesize highly structured faces in Fig.~\ref{fig:exp-pggan-celeba}, which provides convincing evidence that the convolutional generators depend on the implicit positional encoding to obtain spatial information.

To further verify that the lack of positional information causes the higher SWD, we introduce sinusoidal positional encoding (SPE) to the padding-free PGGAN and DCGAN.
The SPE is only added with the input feature map ahead of the convolutional generators.
As shown in Tab.~\ref{tab:pggan-res}(c) and Tab.~\ref{tab:dcgan-lsun}(c), the explicit positional encoding mitigates the effects of removing implicit spatial inductive bias in each convolutional layer. 
Thanks to the explicit positional encoding, the synthesized images in Fig.~\ref{fig:exp-pggan-celeba}(c) can recover the faithful spatial structure.
More results and implementation details are presented in our supplementary material.

\begin{table*}[tb]
    \setlength{\abovecaptionskip}{2pt}
    \centering
    \small
    \caption{Main results for our MS-PIE with $ 256^2$ StyleGAN2 in the FFHQ dataset. The precision and recall are calculated at the same scale as the Fr\'{e}chet inception distance (FID).`C2' indicates the channel multiplier in the generator is two.}
    \begin{tabular}{c|l|c|cccc|cccc}
    \bottomrule
    \multicolumn{2}{l|}{\multirow{2}{*}{\textbf{Training configuration}}}                                                                  & \multirow{2}{*}{Resize} & \multicolumn{2}{c}{FID@512$\downarrow$} & \multirow{2}{*}{\begin{tabular}[c]{@{}c@{}}Precision\\ (\%)$\uparrow$\end{tabular}} & \multirow{2}{*}{\begin{tabular}[c]{@{}c@{}}Recall\\ (\%)$\uparrow$\end{tabular}} & \multicolumn{2}{c}{FID@256$\downarrow$} & \multirow{2}{*}{\begin{tabular}[c]{@{}c@{}}Precision\\ (\%)$\uparrow$\end{tabular}} & \multirow{2}{*}{\begin{tabular}[c]{@{}c@{}}Recall\\ (\%)$\uparrow$\end{tabular}} \\
    \multicolumn{2}{l|}{}                                                                                                         &                         & 20M          & 25M          &                                                                           &                                                                        & 20M          & 25M          &                                                                           &                                                                        \\ \hline \hline
    \multicolumn{2}{l|}{(a) StyleGAN2-C2-256}  & --- & --- & --- & --- & ---- & 5.62 & 5.56 & 75.92 & 51.24  \\ 
    \multicolumn{2}{l|}{(b) StyleGAN2-C2-512}  & --- & 3.47 & 3.41 &75.88 & 54.61 & 5.00 & 4.91 & 75.65 & 54.58 \\ \hline
    \multirow{5}{*}{\begin{tabular}[c]{@{}c@{}}MS-PIE\\ w/\\ padding\end{tabular}}  & (c) Leanable constant input                             & Interp    & 3.46  & 3.35 & 73.84 & 55.77 & 4.82  & 4.50 &  72.75  & 55.42 \\
                                                                           & (d) Cartesian spatial grid & Interp & 3.59  & 3.50 & 73.28 & 56.16 &  \textbf{4.74}   & 4.71 &  73.34 & 55.07 \\ \cline{2-3}
                                                                           & (e) SPE-interp & Interp  & 3.41  & 3.15 & \textbf{74.13} &  56.88 & 4.99 & 4.73 & 73.28 & \textbf{56.93}    \\ \cline{3-3}
                                                                           &  (f)\, SPE-expand  & Expand   & \textbf{3.31}  & \textbf{2.93}  & 73.51 & \textbf{57.32}  & 4.79  & \textbf{4.27} & \textbf{73.48}  & 55.69  \\ \cline{2-11}
                & (g) SPE-expand-C1  &  Expand & 3.65  & 3.40  & 73.05 & 56.45  & 5.54  & 4.83 & 73.59 &  54.29   \\ \hline
    \multirow{4}{*}{\begin{tabular}[c]{@{}c@{}}MS-PIE\\w/o\\ padding\end{tabular}} & (h) Learnable constant input & Interp & 4.99 & 4.01 &  72.81 & 54.35 &  5.67  & 5.11 & 72.37 & 55.52 \\
                                                                           & (i)\, Cartesian spatial grid     & Interp  & \textbf{3.96}  & \textbf{3.76} & \textbf{73.26} &  \textbf{54.71}   & \textbf{5.30}  & \textbf{5.09} & 70.74 & 56.06  \\ \cline{2-3}
                                                                           & (j)\, SPE-interp & Interp  & 4.80 & 4.23 & 73.11 & 54.63 &  5.89 & 5.38 &71.21& \textbf{56.21} \\
                                                                           & (k) SPE-expand & Expand   & 4.46   & 4.17 & 73.05 & 51.07 & 6.08 & 5.59 & \textbf{72.65} & 49.74 \\ \toprule 
    \end{tabular}
    \label{tab:ms-exp}
    \vspace{-8pt}
\end{table*}

\subsection{MS-PIE in StyleGAN2}
\label{sec:exp-ms}
In Tab.~\ref{tab:ms-exp}, we examine our MS-PIE in $256^2$ StyleGAN2-C2 with multiple image scales of $256^2$, $384^2$ and $512^2$.
For the $512^2$ StyleGAN2 baseline in Tab.~\ref{tab:ms-exp}(b), we also downsample the generated images to obtain the FID and P\&R at $256^2$ scale as another strong baseline.   

With zero padding holding spatial anchors, the SPE with `Expand' resizing mode offers a stable explicit transformation between locations which is unchanged during the multi-scale synthesis.
%
As shown in Tab.~\ref{tab:ms-exp}(f), even if containing fewer upsampling blocks and seeing fewer images at $512^2$ scale, the $256^2$ StyleGAN2 can still achieve impressive improvement in both FID and P\&R.
However, once the transformation is changed with the multiple input scales like Tab.~\ref{tab:ms-exp}(d) and Tab.~\ref{tab:ms-exp}(e), there will be a decline in the generation quality at each scale.
This phenomenon indicates that the scale-variant transformation between positional encodings prevents the model from easily sharing the learned information among different scales.

\begin{figure}[tb]
    \setlength{\abovecaptionskip}{5pt}
    \centering
    \includegraphics[width=\linewidth]{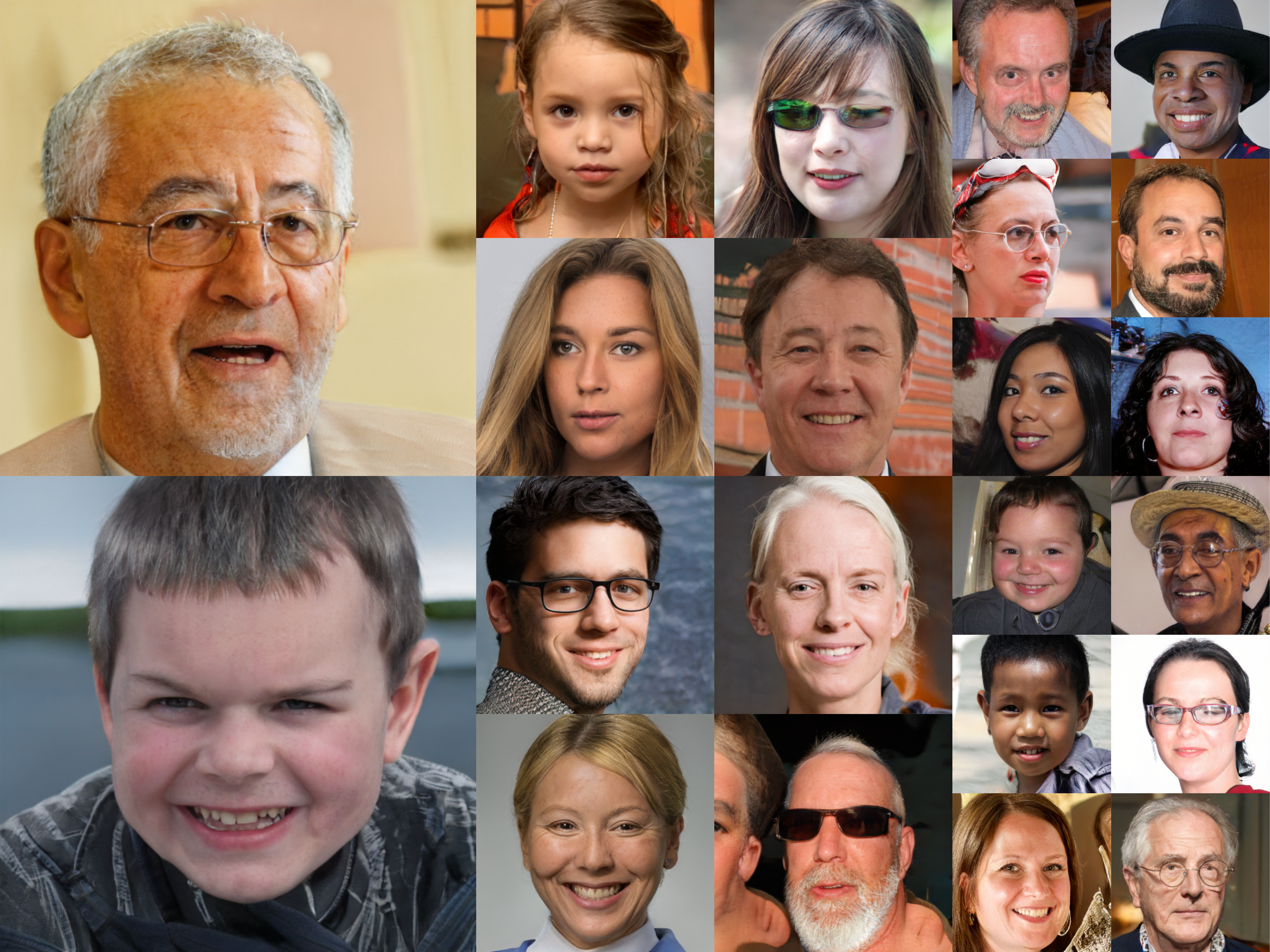}
    \caption{Multi-scale results from configuration (f) in Tab.~\ref{tab:ms-exp}. The results are in $512^2$, $384^2$, and $256^2$ resolution. The higher resolution is presented with a larger size. Please see our appendix and supplementary video for more results.}
    \label{fig:ms-exp-512}
    \vspace{-18pt}
\end{figure}

\begin{figure*}[tb]
    \setlength{\abovecaptionskip}{3pt}
    \centering
    \includegraphics[width=\linewidth]{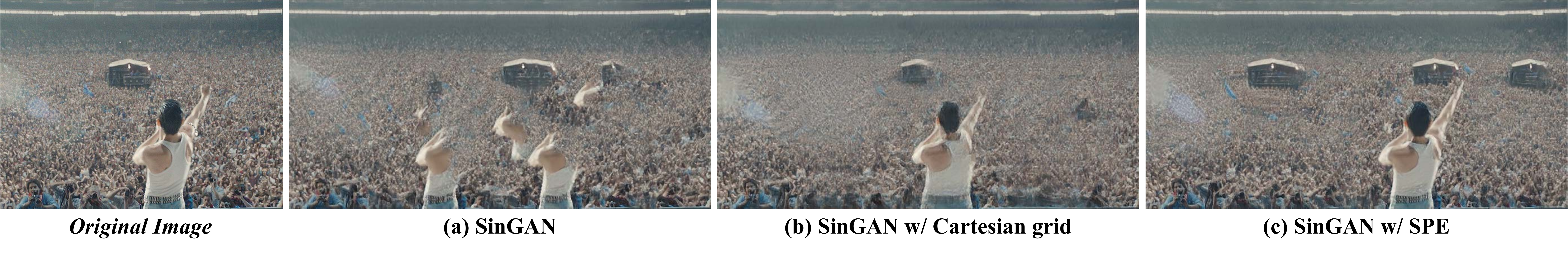}
    \caption{Results of SinGAN with different positional embedding strategies. The original image ($370\times 500$) is taken from the movie `\textit{Bohemian Rhapsody}' and the sampled images are 1.5 wider than the original image.}
    \label{fig:bohemian-singan}
    \vspace{-10pt}
\end{figure*}

%
\noindent \textbf{Without Padding.}
Due to the dynamic training scales in our MS-PIE, the spatial anchors are essential for providing unique reference points to directly align the global structure across different image spaces. 
As shown in Tab~\ref{tab:ms-exp}(h)-(k), abandoning the implicit spatial anchors in each convolutional layer causes an inferior generation quality. 
%
However, to some extent, spatial anchors in the Cartesian spatial grid guide the padding-free generator to retain a faithful global structure at multiple scales.
%
Therefore, containing explicit spatial anchors, Tab.~\ref{tab:ms-exp}(i) achieves the least decline in performance among the other positional encodings.

\noindent \textbf{Lite Model.}
As shown in Tab.~\ref{tab:ms-exp}(g), we select the best configuration (f) and reduce its channel multiplier to investigate the influence of architecture design.
Even if we reduce half of the channels in some layers, our MS-PIE enables the lite generator to yield comparable generation quality to the heavy baseline model.

\noindent \textbf{Qualitative Results.}
Figure~\ref{fig:ms-exp-512} presents the multi-scale images generated from the best training configuration (f) in Tab.~\ref{tab:ms-exp}.
The diverse and realistic multi-scale synthesis further demonstrates the effectiveness of our MS-PIE and the impact of introducing appropriate positional encoding.
Importantly, with the help of MS-PIE, the StyleGAN2 that is originally designed for $256^2$ generation can also synthesize images in more challenging resolutions, like $896^2$ and $1024^2$.
More training configurations and detailed results in higher resolutions can be found in Appx.~\ref{appx:ms}.

As for image manipulation~\cite{abdal2020styleflow}, we customize a convenient pipeline by improving the closed-form factorization~\cite{shen2020closed}.
This indicates that our MS-PIE constructs a shortcut for high-resolution image manipulation with a single backbone.
The implementation details and high-quality manipulation results are shown in Appx.~\ref{sec:appx-manipulation}.

\vspace{-3pt}
\subsection{SinGAN with Positional Encoding}
\vspace{-3pt}
This section demonstrates the effectiveness of explicit positional encoding in SinGAN with a challenging case.
In Fig.~\ref{fig:bohemian-singan}, we take a picture from the famous movie `\textit{Bohemian Rhapsody}', where Mercury is singing to thousands of audiences.
We use SinGAN to extrapolate the original scene so that Mercy can meet more audiences in a more spacious gym.
However, with zero padding offering unbalanced spatial inductive bias, SinGAN can only capture the detailed structure at the borders of the image but generate highly unstructured contents at the center of Fig~\ref{fig:bohemian-singan}(a). 

Different explicit spatial inductive bias enables SinGAN to fulfill various requirements.
Adopting the position-aligned noise map with the Cartesian grid, SinGAN can easily interpolate the coarse structure to a larger size, as shown in Fig~\ref{fig:bohemian-singan}(b).
It is the spatial anchors that keep the positions of major contents (Mercury and the tent) unchanged.
Meanwhile, due to the scale-variant transformation in Eq.~\eqref{eq:trans-csg}, the detailed spatial structure will be stretched, \eg, Mercury accidentally gains a lot of weight.
On the contrary, the scale-invariant spatial inductive bias in SPE leads to a more reasonable patch recurrence in Fig.~\ref{fig:bohemian-singan}(c), while it cannot avoid the shift in the position of Mercury. 
More results about the versatile image manipulation with positional encoding are presented in Appx.~\ref{appx:singan}.

\vspace{-8pt}
\section{Conclusion}
\vspace{-3pt}

In this work, we have a thorough study on how zero padding accidentally encodes imperfect spatial bias for convolutional generators. We have also discussed the strengths of introducing explicit positional encodings, including CSG and SPE, in various existing generator architectures.
%
%
With the flexible explicit positional encoding, we propose a new multi-scale training strategy (MS-PIE) to achieve high-quality image synthesis at multiple scales with a single $256^2$ StyleGAN2.
%
%
We further show that adopting explicit positional encoding can improve the versatility and robustness of SinGAN.


{\small
\bibliographystyle{ieee_fullname}
\bibliography{egbib}
}



\onecolumn
\appendix

\section{Padding Encodes Spatial Bias for Convolutional Generators}
\label{appx:translation-invariance}
In Appx.~\ref{sec:appx-weak} and Appx.~\ref{sec:appx-leak}, we will provide the detailed deviation for Sec.~\ref{sec:translation-invariance} and Sec.~\ref{sec:mechanism-pad}, respectively.
In addition, the effects of the padding mode will also be analyzed with an interesting example in Appx.~\ref{sec:appx-leak}. 
We show that the spatial consistent nonlinear activation function cannot influence the weak stationarity in Appx.~\ref{sec:appx-nonlinear}. 
Section \ref{sec:appx-bias} shows the bias term will not influence the weak stationarity in the convolutional feature map.
Finally, we discuss zero padding from another view by presenting the effects of zero padding in different convolutional layers.

\subsection{Detailed Proof for Weak Stationarity}
\label{sec:appx-weak}
Following the idea in Sec.~\ref{sec:translation-invariance}, we will first consider the effects of translation invariance in the padding-free convolutional generators and give some preliminaries in the stochastic process.
Then, with zero padding, the expectation ($\mathbb{E}$) and autocorrelation function ($R$) for the features shows how padding leaks spatial information.

In this work, we mainly focus on the standard convolution layer with the nonlinear activation layer and take the commonly used LeakyReLU (Eq.~\eqref{eq:leakyrelu}) function as an example. 
The batch normalization and instance normalization are spatial identical operation and can be merged into convolutional operation~\cite{qiao2019weight,karras2020analyzing}. Therefore, we will neglect these layers in the following proof.
\begin{equation}
    LeakyReLU(y) = 
    \begin{cases}
        y, & y\geq 0 \\
        \gamma y, & otherwise
    \end{cases}.
    \label{eq:leakyrelu}
\end{equation}

Taking the spatial noise map ($X_{\vec{\mathbf{i}}}\stackrel{i.i.d.}{\sim} \mathcal{N}(0,1)$) as input, the expectation of the first convolutional feature map ($\mathbb{E}(y_{\vec{\mathbf{i}}}^{(1)})$) is:
%
\begin{align}
    \mathbb{E}(y_{\vec{\mathbf{i}}}^{(1)}) & = \sum_{k} w_k^{(1)}\int_{-\infty}^{+\infty}x_kp(x_k)dx_k + b^{(1)} \notag \\
                            \label{eq:appx-eq-e1}
                            & = \sum_k w_k^{(1)} \mathbb{E}(x_k) + b^{(1)} \\
                            \label{eq:appx-eq-e2}
                            & = b^{(1)}.
\end{align}
Here, as we assume the input is sampled from a zero-expectation distribution, the final results in Eq.~\eqref{eq:appx-eq-e2} shows the expectation is only related to the bias parameters in convolutional layers.
However, this is not the general formulation for the expectation of the feature map in the convolutional generators.
%
%

%
After adopting a LeakyReLU function ($g$) and the next convolutional layer, we will obtain the general formulation of $\mathbb{E}(y_{\vec{\mathbf{i}}})$:
\begin{align}
    \mathbb{E}(y_{\vec{\mathbf{i}}}^{(2)}) &= \sum_{k} w_k\int_{-\infty}^{+\infty} g(y^{(1)}_k) p(y^{(1)}_k)dy^{(1)}_k + b^{(2)} \notag \\
    &= \sum_k w_k^{(2)}\int_{-\infty}^{0} \gamma y^{(1)}_k p(y^{(1)}_k) dy^{(1)}_k  
    + \sum_k w_k^{(2)} \int_0^{+\infty} y^{(1)}_k p(y^{(1)}_k) dy^{(1)}_k + b^{(2)} \notag \\
    \label{eq:appx-nopad}
    &= \sum_k w_k^{(2)} \cdot (\gamma\mathbb{C}_1 + \mathbb{C}_2) + b^{(2)},
\end{align}
where $\mathbb{C}_1, \mathbb{C}_2$ are constants from the piecewise-defined finite integration. Therefore, the translation-invariant convolutional layer results in a spatial consistent expectation of the feature map.
The expectation is a linear combination of the parameters in the convolution kernel, which is irrelevant to the positions.

\begin{figure}[tb]
    \centering
    \small
    \includegraphics[width=0.7\linewidth]{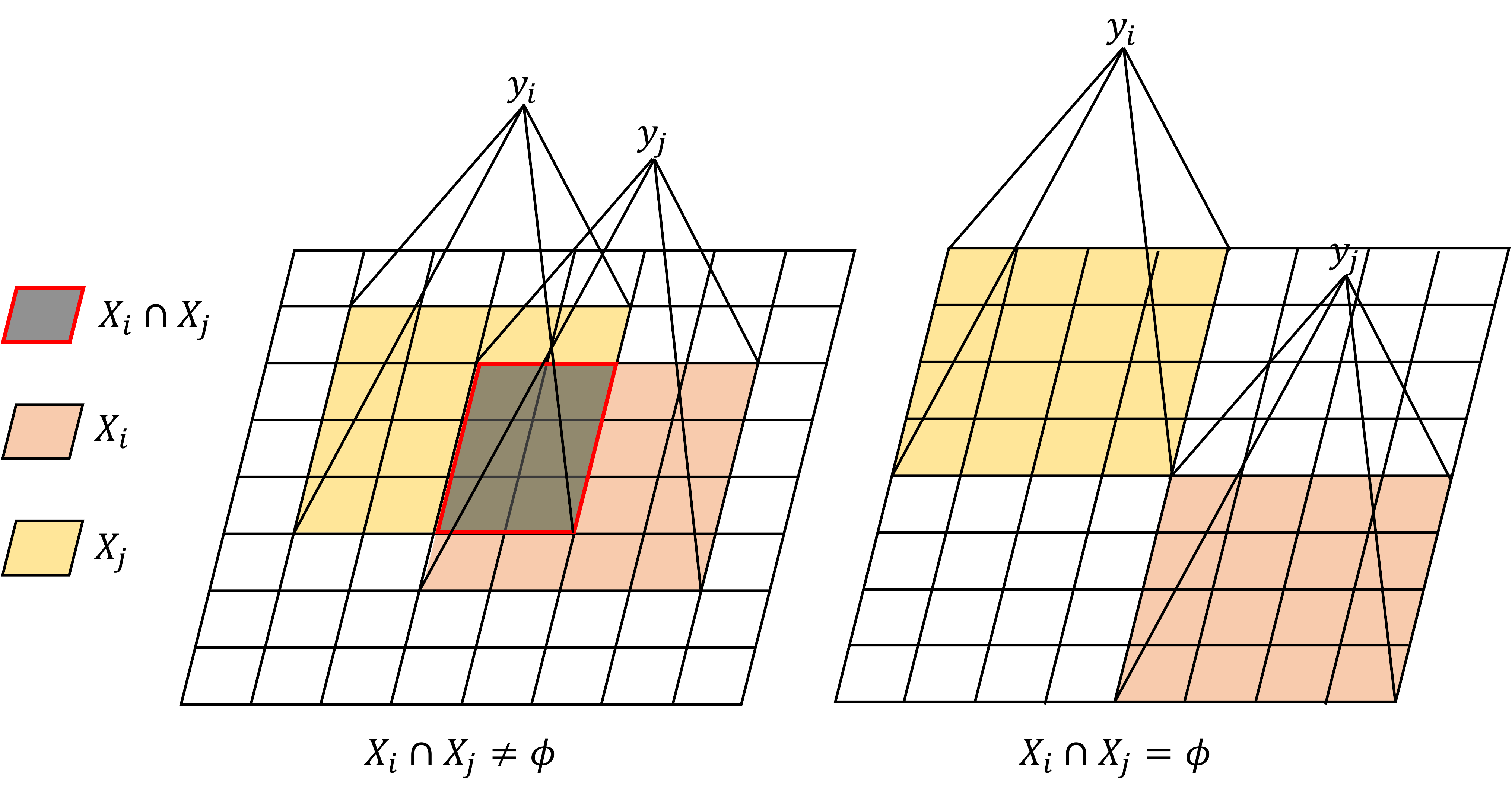}
    \caption{An illustration for the two cases of calculating the autocorrelation function in the convolutional feature map.}
    \label{fig:appx-r-proof}
\end{figure}

$R(y_{\vec{\mathbf{i}}}, y_{\vec{\mathbf{j}}})$ portrays the relationship between two spatial locations. 
Here, taking $X_{\vec{\mathbf{i}}}\stackrel{i.i.d.}{\sim} \mathcal{N}(0,1)$, we directly analyze the feature map after the convolutional operation. The bias item is removed in this part since we do not care about the spatial consistent addition component in autocorrelation analysis.
However, as shown in Fig.~\ref{fig:appx-r-proof}, we should consider two cases of whether the input patch regions ($X_{\vec{\mathbf{i}}}, X_{\vec{\mathbf{j}}}$) are seperated.
If $X_{\vec{\mathbf{i}}}, X_{\vec{\mathbf{j}}}$ are separated, it is trivial to calculate$R(y_{\vec{\mathbf{i}}}, y_{\vec{\mathbf{j}}})$: 
\begin{align}
    R(y_{\vec{\mathbf{i}}}, y_{\vec{\mathbf{j}}})  &= \mathbb{E}(y_{\vec{\mathbf{i}}}y_{\vec{\mathbf{j}}})      \notag \\
                 &= \mathbb{E}[(\sum_kw_kx_k)(\sum_tw_tx_t)]  \notag \\
                 &= \sum_{k,t} w_k w_t \mathbb{E}(x_k) \mathbb{E}(x_t) \notag \\
                 \label{eq:appx-r0}
                 &= 0
\end{align}

Once there lies an intersection between the two input features, the autocorrelation function $R(y_{\vec{\mathbf{i}}}, y_{\vec{\mathbf{i}}})$ should be:
\begin{align}
    R(y_{\vec{\mathbf{i}}}, y_{\vec{\mathbf{j}}}) &= \mathbb{E}(y_{\vec{\mathbf{i}}}, y_{\vec{\mathbf{j}}})      \notag \\
                &= \mathbb{E}[(\sum_kw_kx_k)(\sum_tw_tx_t)]  \notag \\
                \label{eq:appx-2parts}
                &= \mathbb{E}[\sum_{x_l \in X_i \cap X_j} w_{k_l}w_{t_l} x_l^2]
                 + \mathbb{E}[\sum_{p \neq q}  w_{k_p}w_{t_q} x_px_q] \\
                \label{eq:appx-r1}
                &= \sum_{x_l \in X_i \cap X_j}  w_{k_l}w_{t_l} \mathbb{E}(x_l^2)  \\
                &= R(\vec{\mathbf{i}} - \vec{\mathbf{j}}),
\end{align}
where $p\neq q$ indicates that $x_p$ and $x_q$ are not the same feature variable.
As shown in Eq.~\eqref{eq:appx-2parts}, the formulation can be split into two parts of the intersection part and the separated part.
The spatial independence causes the separated part to be zero.
The intersection part is related to the variance of the random variables and the parameters in the convolutional kernel.
As the input $X$ is spatial identical, $E(x_l^2)$ must be consistent among the spatial space.
Thus, Eq.~\eqref{eq:appx-r1} is only related the intersection mode of $X_{\vec{\mathbf{i}}} \cap X_{\vec{\mathbf{i}}}$, which is determined by $\vec{\mathbf{i}} - \vec{\mathbf{j}}$.
In summary, after fusing Eq.~\eqref{eq:appx-r0} and Eq.~\eqref{eq:appx-r1}, the autocorrelation function of a feature after convolution operation should be:
\begin{align}
    R(y_{\vec{\mathbf{i}}}, y_{\vec{\mathbf{j}}}) &= R(\vec{\mathbf{i}} - \vec{\mathbf{j}}) \notag \\
                \label{eq:appx-r-nopad}
                &= \sum\limits_{x_l \in X_{\vec{\mathbf{i}}} \cap X_{\vec{\mathbf{j}}}} w_{k_l}w_{t_l}  \mathbb{E}(x_l^2).
\end{align}
%
Here, a large offest vector ($\vec{\mathbf{i}} - \vec{\mathbf{j}}$) will result in no intersection between two input feature maps. 
Then, there will not exist any element ($x_l$) belonging to $X_{\vec{\mathbf{i}}} \cap X_{\vec{\mathbf{j}}}$  and the autocorrelation function will be zero as Eq.~\eqref{eq:appx-r0}.

\subsection{How Padding Serves as Spatial Bias}
\label{sec:appx-leak}

Taking the zero padding into consideration, the linear combination of the parameters in the convolutional kernels in Eq.~\eqref{eq:appx-nopad} and Eq.~\eqref{eq:appx-r-nopad} will be influenced:
\begin{align}
    \mathbb{E}(y_{\vec{\mathbf{i}}}^{(2)})  &= \sum_{k} w_k\int_{-\infty}^{+\infty} g(y^{(1)}_k) p(y^{(1)}_k)dy^{(1)}_k \cdot \mathbbm{1}(y_k^{(1)}\notin Pad)+ b^{(2)} \notag \\
    \label{eq:appx-e-pad}
        &= \sum_k w_k^{(2)} (\gamma\mathbb{C}_1 + \mathbb{C}_2) \mathbbm{1}(y_k^{(1)}\notin Pad) + b^{(2)},
\end{align}
%
%
\begin{align}
    R(y_{\vec{\mathbf{i}}}, y_{\vec{\mathbf{j}}})
                &= \sum\limits_{x_l \in X_{\vec{\mathbf{i}}} \cap X_{\vec{\mathbf{j}}}} w_{k_l}w_{t_l}  \mathbb{E}(x_l^2) \cdot \mathbbm{1}(x_l \notin Pad) \\
                &\neq R(\vec{\mathbf{i}} - \vec{\mathbf{j}}),  
\end{align}
where the indicator function $\mathbbm{1}(x_i \notin Pad)$ determines whether current input belongs to zero padding regions.

Here, we further investigate the effects of padding modes on the spatial information leak. 
The effects of the padding mode can also be clarified from the view of the expectation and autocorrelation function.
Reflection padding will be taken as an example in the following analysis.
In such padding mode, the boundary variables are copied to the padding regions.
When analyzing the expectation ($\mathbb{E}(y_{\vec{\mathbf{i}}}$), we find the linear combination will not be influenced as Eq.~\eqref{eq:appx-e-pad}.
The expectation with reflection padding mode is the same as the padding-free formulation in Eq.~\eqref{eq:appx-nopad}.
However, the autocorrelation function ($R(y_{\vec{\mathbf{i}}}, y_{\vec{\mathbf{i}}})$) is influenced by the reflected padding:
\begin{align}
    R(y_{\vec{\mathbf{i}}}, y_{\vec{\mathbf{j}}})
                &= \mathbb{E}[(\sum_kw_kx_k)(\sum_tw_tx_t)]  \notag \\
                &= \mathbb{E}[\sum_{x_l \in X_{\vec{\mathbf{i}}} \cap X_{\vec{\mathbf{j}}}} w_{k_l}w_{t_l} x_l^2]
                 + \mathbb{E}[\sum_{p \neq q}  w_{k_p}w_{t_q} x_px_q] \\
                \label{eq:appx-r-reflect}
                &= \sum\limits_{x_l \in X_{\vec{\mathbf{i}}} \cap X_{\vec{\mathbf{j}}}} w_{k_l}w_{t_l} \mathbb{E}(x_l^2)
                   + \sum_{p \neq q} w_{k_p}w_{t_q} \mathbb{E}(x_p^2)
                   \mathbbm{1}(x_p \in Pad, x_p = x_q ) \\
                &\neq R(\vec{\mathbf{i}} - \vec{\mathbf{j}}),  
\end{align}
\begin{wrapfigure}{l}{0.5\textwidth}
    \centering
    \includegraphics[width=0.45\textwidth]{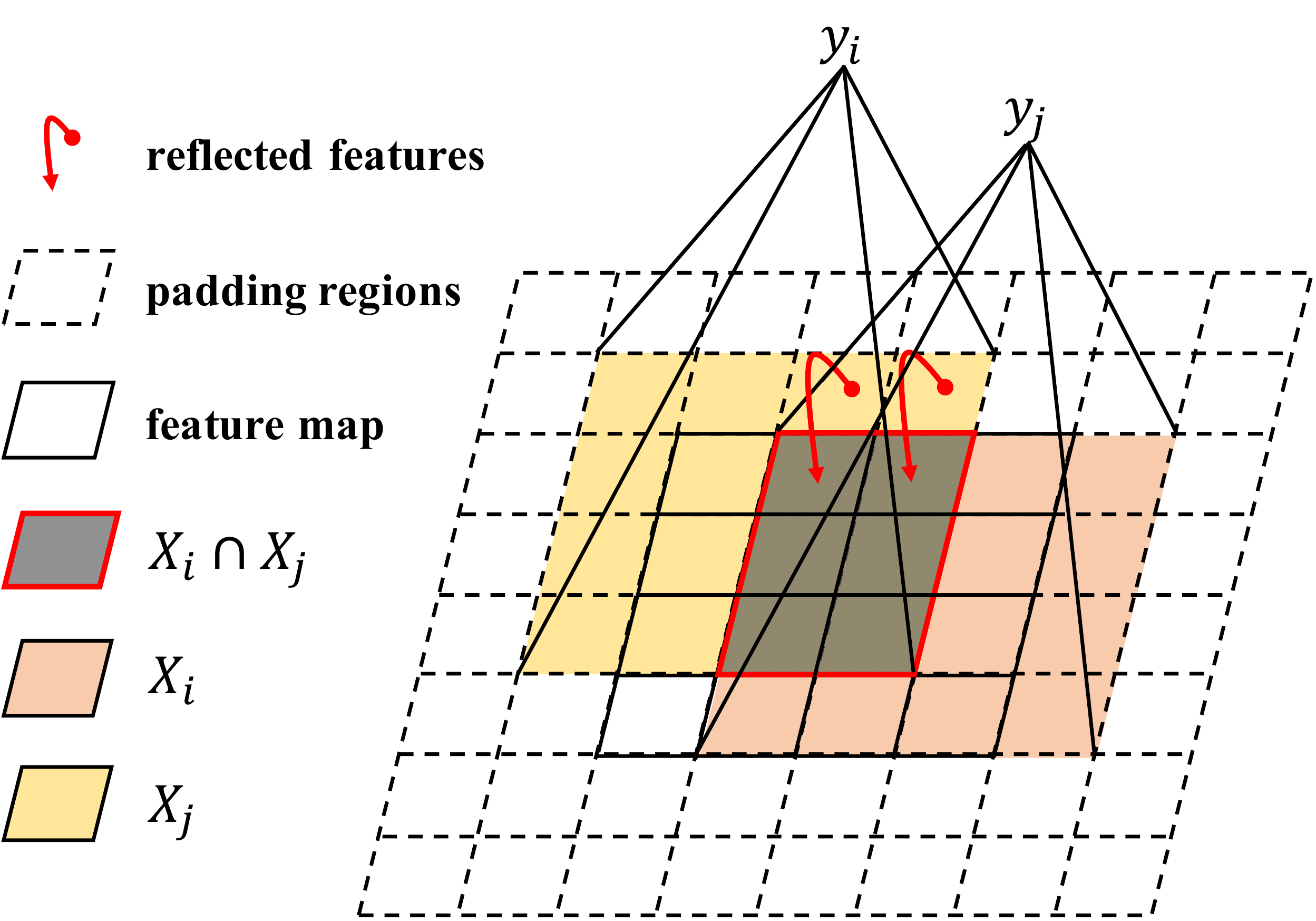}
    \caption{Illustration for refelcted padding.}
    \label{fig:appx-reflect}
\end{wrapfigure}
where $\mathbbm{1}(x_p \in Pad, x_p = x_q)$ indicates the whether $x_p$ belongs to the padding regions and $x_q$ is the corresponding boundary variable copied by $x_p$. 
As shown in Fig.~\ref{fig:appx-reflect}, the red arrow presents the case in $\mathbbm{1}(x_p \in Pad, x_p = x_q)$.
Unlike the zero padding, the reflection padding can introduce extra relations in the non-intersection regions.
As shown in Eq.~\eqref{eq:appx-r-reflect}, though the reflected padding does not change the expectation distribution, it adds extra items related to the padding in the autocorrelation function ($R(y_{\vec{\mathbf{i}}}, y_{\vec{\mathbf{j}}})$). 
Therefore, the reflection padding can inject implicit positional embedding by only changing the distribution of the autocorrelation function, leading to a location-aware autocorrelation function.
Other padding modes like `Circular Padding' in PyTorch can also be analyzed from the view of the expectation and autocorrelation function similarly with Eq.~\eqref{eq:appx-r-reflect}.
In `Circular Padding', the padded variable are borrowed from the other side of the feature map.
Intuitively, the padded feature variables from remote areas can be assumed to be independent to the boundary feature variables. 
Thus, such padding mode cannot encode spatial information for the convolutional networks.

\subsection{Effects of Nonlinear Activation Function on Autocorrelation Function}
\label{sec:appx-nonlinear}

%
Intuitively, as a spatial identical operation, the nonlinear activation function changed the range of each feature, while does not influence the weak stationarity. 
Here, we adopt the LeakyReLU function in Eq.~\eqref{eq:appx-r-nopad} to verify it will not change the weak stationarity.

\begin{align}
    R(g(y_{\vec{\mathbf{i}}}), g(y_{\vec{\mathbf{j}}})) &= \mathbb{E}(g(y_{\vec{\mathbf{i}}})g(y_{\vec{\mathbf{j}}})) \notag \\ 
                     \label{eq:appx-nonlinear-1}
                      &= \int_{-\infty}^{+\infty} \int_{-\infty}^{+\infty} g(y_{\vec{\mathbf{i}}}) g(y_{\vec{\mathbf{j}}}) p(y_{\vec{\mathbf{i}}}, y_{\vec{\mathbf{j}}}) dy_{\vec{\mathbf{i}}}dy_{\vec{\mathbf{j}}} \notag \\
                      &= \int_{-\infty}^{0} \int_{-\infty}^{0} \gamma^2 y_{\vec{\mathbf{i}}} y_{\vec{\mathbf{j}}} p(y_{\vec{\mathbf{i}}}, y_{\vec{\mathbf{j}}}) dy_{\vec{\mathbf{i}}}dy_{\vec{\mathbf{j}}}  + \int_{-\infty}^{0} \int_{0}^{+\infty} \gamma y_{\vec{\mathbf{i}}} y_{\vec{\mathbf{j}}} p(y_{\vec{\mathbf{i}}}, y_{\vec{\mathbf{j}}}) dy_{\vec{\mathbf{i}}}dy_{\vec{\mathbf{j}}}          \notag \\
                      &\quad + \int_{0}^{+\infty} \int_{-\infty}^{0} \gamma y_{\vec{\mathbf{i}}} y_{\vec{\mathbf{j}}} p(y_{\vec{\mathbf{i}}}, y_{\vec{\mathbf{j}}}) dy_{\vec{\mathbf{i}}}dy_{\vec{\mathbf{j}}}   + \int_{0}^{+\infty} \int_{0}^{+\infty} y_{\vec{\mathbf{i}}} y_{\vec{\mathbf{j}}} p(y_{\vec{\mathbf{i}}}, y_{\vec{\mathbf{j}}}) dy_{\vec{\mathbf{i}}}dy_{\vec{\mathbf{j}}}   \\
                      &= F_1(\vec{\mathbf{i}} - \vec{\mathbf{j}}) + F_2(\vec{\mathbf{i}} - \vec{\mathbf{j}}) + F_3(\vec{\mathbf{i}} - \vec{\mathbf{j}}) + F_4(\vec{\mathbf{i}} - \vec{\mathbf{j}}) \\
                      &= R(\vec{\mathbf{i}} - \vec{\mathbf{j}}),
\end{align}
where $F_i(\vec{\mathbf{i}}-\vec{\mathbf{j}})$ denotes the function related to the offet vector $\vec{\mathbf{i}} - \vec{\mathbf{j}}$.
In Eq.~\eqref{eq:appx-nonlinear-1}, the manipulation item ($y_{\vec{\mathbf{i}}} y_{\vec{\mathbf{j}}} p(y_{\vec{\mathbf{i}}}, y_{\vec{\mathbf{j}}}) dy_{\vec{\mathbf{i}}}dy_{\vec{\mathbf{j}}}$) can be expanded similarly with Eq.~\eqref{eq:appx-2parts} and then the final value of such item will be determined by the offet vector.
Therefore, the spatial identical nonlinear activation function like LeakyReLU cannot change the weak stationarity.

\subsection{Bias Term in Autocorrelation Function}
\label{sec:appx-bias}
In Appx.~\ref{sec:appx-weak} and Appx.~\ref{sec:appx-leak}, we do not condider the bias term in the analysis of the autocorrelation function.
Intuitively, the spatially identical addition operation cannot introduce an extra relationship between feature variables.
Here, we further analyze the effects of the bias term on $R(y_{\vec{\mathbf{i}}}, y_{\vec{\mathbf{j}}})$ in Eq.~\eqref{eq:appx-r-nopad}:
\begin{align}
    R(y_{\vec{\mathbf{i}}}, y_{\vec{\mathbf{j}}}) &= \mathbb{E}((y_{\vec{\mathbf{i}}} + b) (y_{\vec{\mathbf{j}}} + b))      \notag \\
        & = \mathbb{E}(y_{\vec{\mathbf{i}}}y_{\vec{\mathbf{j}}} + b \cdot (y_{\vec{\mathbf{i}}} + y_{\vec{\mathbf{j}}}) + b^2) \notag\\ 
        \label{eq:appx-bias-1}
        & = \mathbb{E}(y_{\vec{\mathbf{i}}}y_{\vec{\mathbf{j}}}) + b \cdot (\mathbb{E}(y_{\vec{\mathbf{i}}}) + \mathbb{E}(y_{\vec{\mathbf{j}}})) + b^2 \\
        & = \mathbb{E}(y_{\vec{\mathbf{i}}}y_{\vec{\mathbf{j}}}) + b^2      
\end{align}
The extra $b^2$ term is not related to the absolute positional information. Thus, the spatially identical addition operation cannot influence the weak stationarity in the convolutional feature map.
Besides, as shown in Eq.~\eqref{eq:appx-bias-1}, the weak stationarity indeed implies the property of spatially consistent expectation in Eq.~\eqref{eq:appx-nopad}.
Equation \eqref{eq:appx-r-nopad} reveals that if there is not any intersection between input features, the output features are independent. After adopting the bias term, such independence is also kept:
\begin{align}
    R(y_{\vec{\mathbf{i}}}, y_{\vec{\mathbf{j}}}) &= \mathbb{E}((y_{\vec{\mathbf{i}}} + b) (y_{\vec{\mathbf{j}}} + b))      \notag \\
        & = \mathbb{E}(y_{\vec{\mathbf{i}}}y_{\vec{\mathbf{j}}}) + b^2 \notag \\
        & = b^2 = \mathbb{E}(y_{\vec{\mathbf{i}}} + b) \mathbb{E}(y_{\vec{\mathbf{j}}} + b).
\end{align}

\subsection{Locally Asymmetric and Symmetric Zero Padding}
\label{sec:appx-asymmetric}


\begin{figure}[htb]
    \centering
    \includegraphics[width=\linewidth]{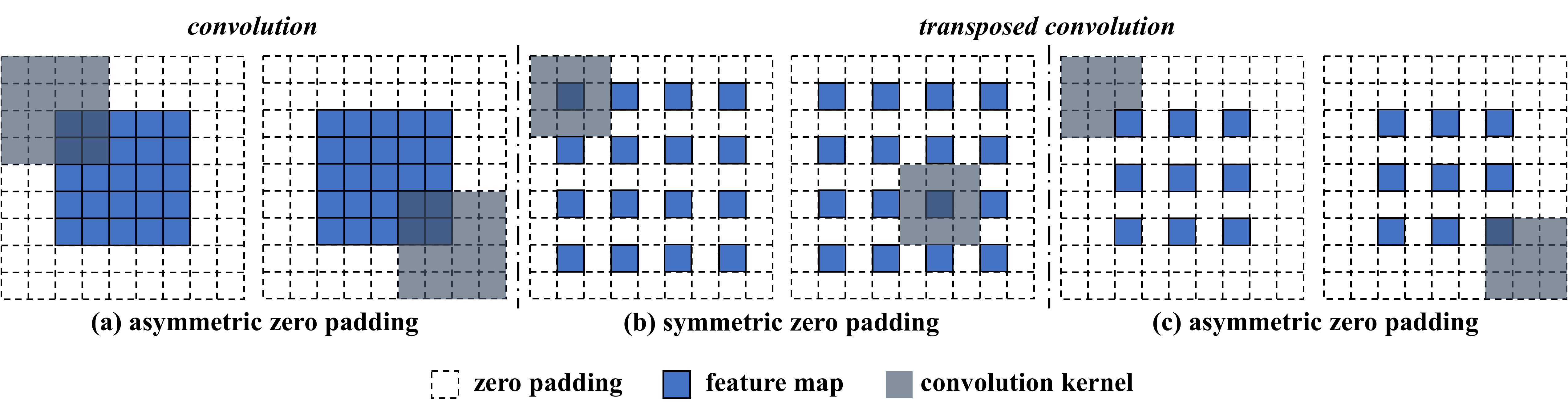}
    \caption{Illustration for asymmetric and symmetric zero padding in the convolution and transposed convolution.}
    \label{fig:appx-asymmetric-pad}
\end{figure}


In this section, we will clarify that it is the asymmetric zero padding that introduces the implicit spatial bias.
Firstly, the asymmetric zero padding denotes the cases shown in Fig.~\ref{fig:appx-asymmetric-pad}(a), where the convolutional kernel covers the locally asymmetric zero padding at corners.
Indeed, zero padding adopted in the convolutional layers is always locally asymmetric, from the view of the effective receptive field.
However, in the transposed convolution layer, Fig.~\ref{fig:appx-asymmetric-pad}(b) presents a case in which the effective zero padding on the input feature is locally symmetric.
As for the transposed convolutional layer for $2\times$ upsampling in generators, the commonly used setting also brings locally asymmetric zero padding, as illustrated in Fig.~\ref{fig:appx-asymmetric-pad}(c).
StyleGAN2 adopts the transposed convolutional layer with the same padding mode as Fig.~\ref{fig:appx-asymmetric-pad}(c).
As analyzed in Sec. ~\ref{sec:mechanism-pad}, the locally symmetric zero padding cannot introduce any spatial bias, because the convolutional kernel will meet the same padding patterns anywhere during its sliding over the feature map. 
On the contrary, Fig.~\ref{fig:appx-asymmetric-pad}(a) and Fig.~\ref{fig:appx-asymmetric-pad}(c) present the asymmetric zero padding pattern at corners which encodes an implicit spatial anchor for the convolutional generators.



\section{StyleGAN2}
\label{appx:stylegan2}
\subsection{Implementation Details}
We directly follow the original setting detailed in \cite{karras2020analyzing} to train the StyleGAN2 model mentioned in our study.
Generally, the channel multiplier is set to two (C2), which indicates the `Large Network' proposed in the original StyleGAN2.
As for the evaluation metric, we compute each metric three times with different random seeds and report their average. The FID is calculated with 50,000 real images while the P\&R is computed with 10,000 real images.

\subsection{Padding Effects on StyleGAN2}
\begin{figure}[htb]
    \includegraphics[width=0.9\linewidth]{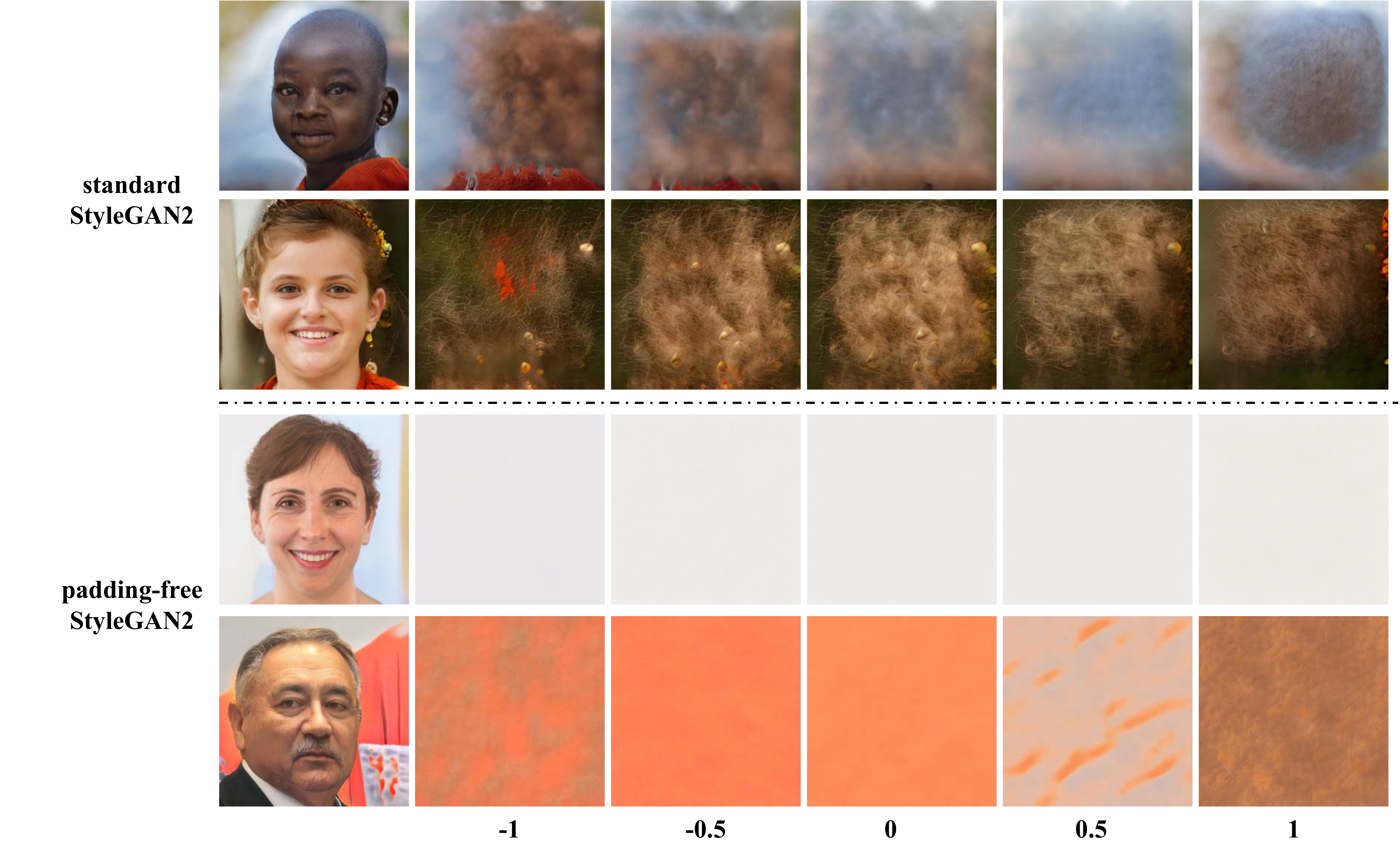}
    \caption{More results for standard StyleGAN2 (above the dotted line) and padding-free StyleGAN2 (under the dotted line). The first column is sampled with the original leaned constant input. The other five columns are sampled with different identical values (from left to right: -1, -0.5, 0, 0.5, 1) filling in the learned constant input at the start of the convolutional generator.}
    \label{fig:appx-pad-effect}
\end{figure}

Section 3.3 describes the experiments on StyleGAN2 for investigating the padding effects.
The motivation of the experiments is to adopt spatially identical input in StyleGAN2 to show the behavior of the convolutional generator. 
In StyleGAN2, there lie three input signals, \ie, style latent code, spatial noise map, and the learnable constant input.
Intuitively, the style latent code is originally identical in spatial space while the spatial noise map is sampled from spatially \iid~noise distribution.
Thus, we only need to modify the learnable constant input at the start of the generator to guarantee that the input signals are spatially identical.
In this experiment, we directly fill in the learned constant input with an identical value.
Figure \ref{fig:appx-pad-effect} presents more results sampled from the standard StyleGAN2 and padding-free StyleGAN2 with different identical input values.

As discussed in Sec.~\ref{sec:translation-invariance}, if the generator is translation-invariant, the spatially identical input signal should have resulted in spatially identical colors or patterns in the output images.
However, the standard StyleGAN2 with a fully-convolutional generator in Fig.~\ref{fig:appx-pad-effect} presents a biased spatial structure in which the borders are fixed to a frozen pattern.
Once the padding is removed from the generator, the spatially identical colors or patterns will cover the whole canvas, as shown in Fig.~\ref{fig:appx-pad-effect}.
This experiment does not rely on any assumption on the convolutional weights like \cite{alsallakh2020mind}.
Thus, we believe that it is a more reasonable and convenient way to show the padding effects on the convolutional generator.

\section{PGGAN and DCGAN}
\label{appx:pggan}

\subsection{Implementation Details}
We follow the original training configuration in the training of PGGAN. In addition to the experiments on the cropped CelebA dataset, we also present the results on the LSUN Bedroom dataset in Tab.~\ref{tab:appx-pggan-res}.
As for DCGAN, the baseline architecture used in our experiments only contains upsampling layers and convolutional layers, because the zero padding in the transposed convolution layer cannot be removed easily.
Meanwhile, we adopt WGAN-GP~\cite{gulrajani2017improved} in DCGAN to improve the generation quality of the baseline model.
Namely, a much stronger baseline model with DCGAN architecture is used in this study.

\noindent \textbf{Combine with SPE.}
In the experiments on PGGAN and DCGAN, we combine the padding-free generator with SPE to further demonstrate that removing padding directly causes the lack of spatial information.
Since SPE can be constructed with any channel dimension, we combine such flexible explicit positional encoding with the padding-free PGGAN and DCGAN.
SPE is directly added with the first input feature map ahead of the convolutional generator.
To be specified, the first $4 \times 4$ feature map is combined with SPE. Then, such a feature map containing explicit positional information will be adopted as the input of the following convolutional generator.

\begin{table}[htb]
    \centering
    \caption{Multi-level Sliced Wasserstein Distance (SWD)~\cite{karras2017progressive} between the synthesized and training images for different training configurations at $128\times 128$. Each column in SWD represents one level of Laplacian pyramid~\cite{burt1983laplacian}, and the last one offers an average of the three distances. $\downarrow$ indicates lower is better.}
    \begin{tabular}{l|cccc|cccc}
    \bottomrule
    \multirow{3}{*}{\textbf{Training configuration}} & \multicolumn{4}{c|}{\textbf{\textsc{CelebA}}}                              & \multicolumn{4}{c}{\textbf{\textsc{LSUN Bedroom}}}                        \\ \cline{2-9} 
                                            & \multicolumn{4}{c|}{Sliced Wasserstein distance $\times 10^3 \downarrow$} & \multicolumn{4}{c}{Sliced Wasserstein distance $\times10^3 \downarrow$} \\
                                            & 128           & 64          & 32          & Avg.         & 128          & 64          & 32          & Avg.         \\ \hline \hline
    (a) PGGAN                               & \textbf{3.162}& \textbf{4.285}& \textbf{5.000}& \textbf{4.149} & 7.473        & 5.197       & 5.214       & 5.962        \\
    (b) + Remove padding                    & 11.169        & 6.945       & 7.488       & 8.534        & 13.619       & 10.043      & 6.581       & 10.081       \\
    (c) + SPE at head, w/o padding                   & 4.555         & 6.164       & 6.365       & 5.694        & \textbf{6.588}& \textbf{4.437}& \textbf{5.090} & \textbf{5.371}       \\ \toprule
    \end{tabular}
    \label{tab:appx-pggan-res}
\end{table}

\begin{figure}[htb]
    \centering
    \includegraphics[width=\linewidth]{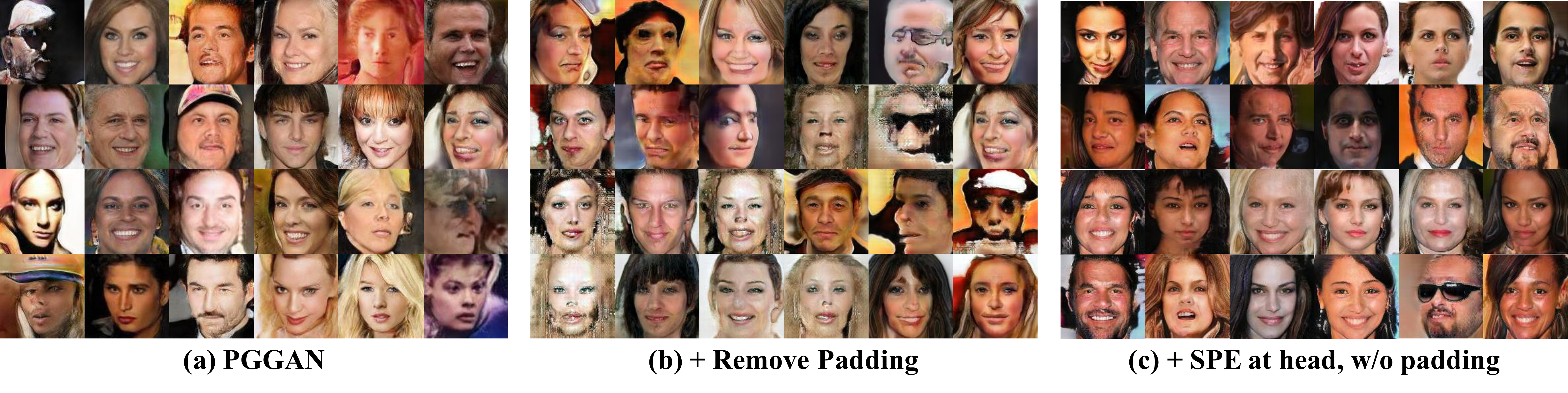}
    \caption{Sampled images from various PGGANs trained on cropped CelebA. (a), (b), and (c) indicate the different training configurations in Tab.~\ref{tab:appx-pggan-res}.}
    \label{fig:appx-pggan-celeba}
\end{figure}

\begin{figure}[htb]
    \centering
    \includegraphics[width=\linewidth]{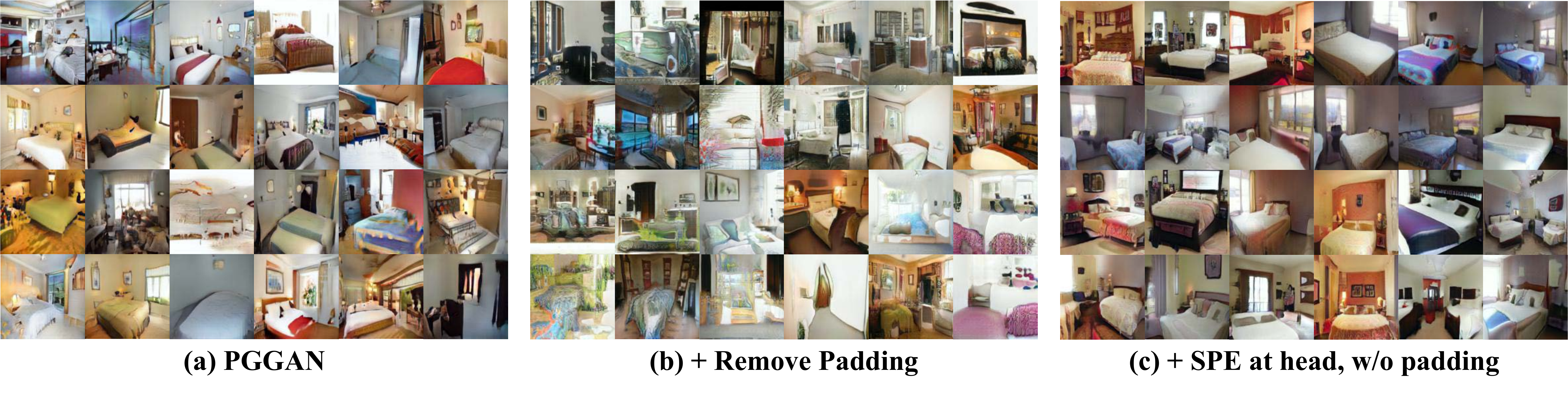}
    \caption{Sampled images from various PGGANs trained on LSUN Bedroom. (a), (b), and (c) indicate the different training configurations in Tab.~\ref{tab:appx-pggan-res}. (\textbf{Best viewed with zoom in.})}
    \label{fig:appx-pggan-lsun}
\end{figure}

\begin{figure}[htb]
    \centering
    \includegraphics[width=\linewidth]{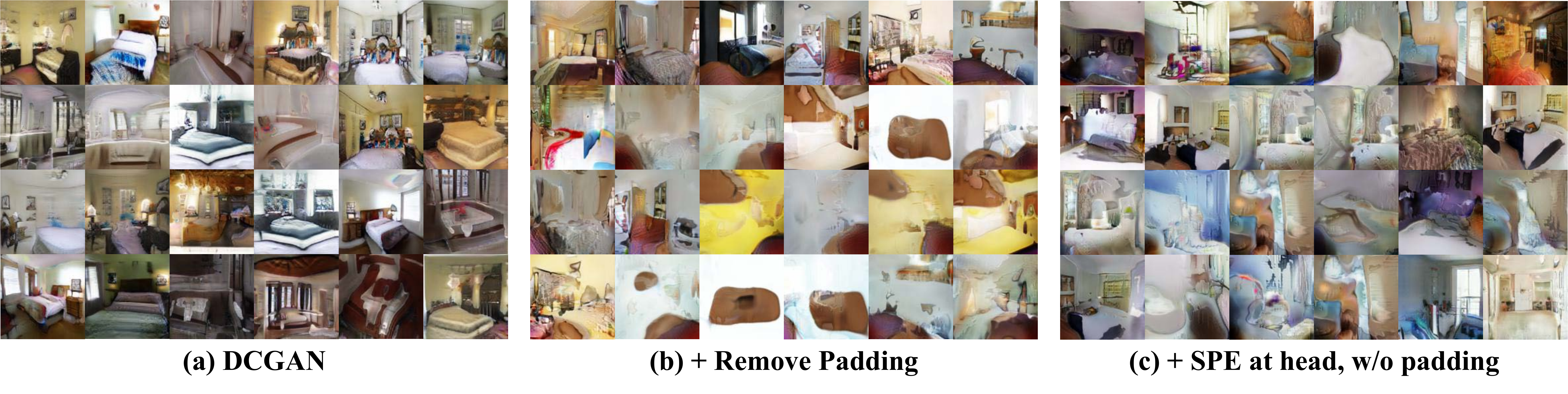}
    \caption{Sampled images from various DCGANs trained on LSUN Bedroom. (a), (b), and (c) indicate the different training configurations in Tab.~\ref{tab:dcgan-lsun}. (\textbf{Best viewed with zoom in.})}
    \label{fig:appx-dcgan-lsun}
\end{figure}

\subsection{More Results and Analyses}
Table \ref{tab:appx-pggan-res} furthter shows the results in LSUN Bedroom dataset with PGGAN.
More qualitative results are presented in Fig.~\ref{fig:appx-pggan-celeba}, Fig.~\ref{fig:appx-pggan-lsun}, and Fig.~\ref{fig:appx-dcgan-lsun}.
In Tab.~\ref{tab:appx-pggan-res}(c), the closer distance in the LSUN Bedroom dataset further demonstrates the impact of the spatial inductive bias to the convolutional generator.  

\section{MS-PIE}
\label{appx:ms}

\subsection{Implementation Details}
In this study, we demonstrate the effectiveness of our MS-PIE in the state-of-the-art $256^2$ StyleGAN2 model that is originally designed for $256\times 256$ image generation.
Three scales will be adopted the multi-scale training strategy while the sampling probability is fixed to $[0.5, 0.25, 0.25]$.
When we replace the learnable constant input with the Cartesian grid, the input feature will only contain two channels.
Other training details strictly follows the original configuration in the StyleGAN2, as mentioned in Appx.~\ref{appx:stylegan2}.

\subsection{MS-PIE in Higher Resolutions}

\begin{figure}[tb]
    \centering
    \includegraphics[width=\linewidth]{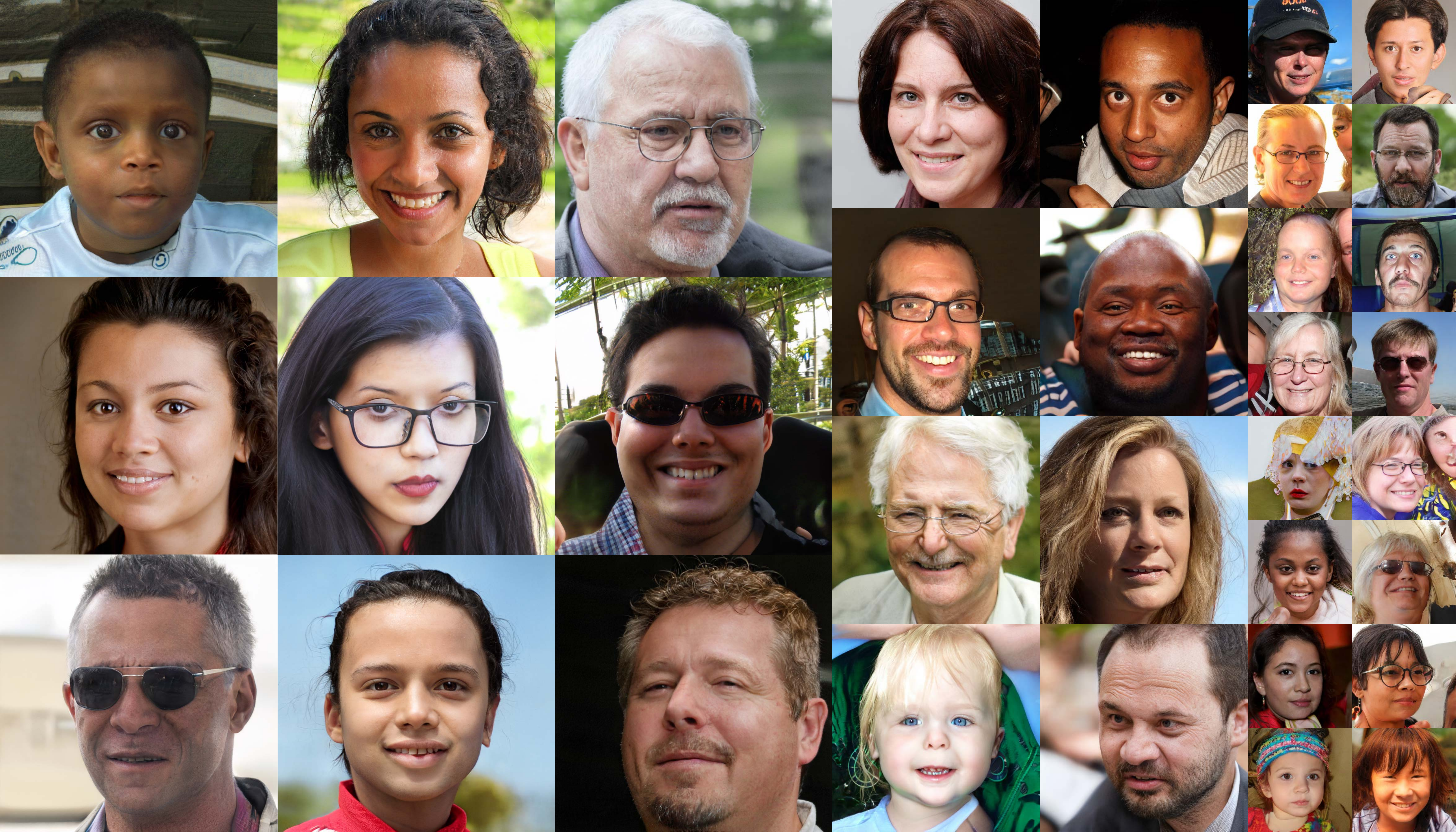}
    \caption{Sampled multi-scale images from $256^2$ StyleGAN2 with MS-PIE. We adopt three scales of $896^2$, $512^2$ and $256^2$ resolutions. The larger resolution is shown with larger size.}
    \label{fig:appx-ms-896}
\end{figure}

With MS-PIE, the original $256^2$ StyleGAN2 containing six upsampling blocks can also achieve high-quality image generation in higher resolutions.
However, we have to admit that our MS-PIE will cause additional GPU memory cost.
This is because current deep learning frameworks, like PyTorch, always store tensors in continuous memory blocks, which will further accelerate the computational speed.
When we switch to a different generation scale from the previous training iteration, we observe that the GPU memory that saves data for the last iteration will not be released or used for the current iteration.
Thus, during training, the total GPU memory cost is not the cost for the largest scale.
The peak of the GPU memory cost may be the sum of the cost for the three different scales. 
We have tested MS-PIE on Nvidia Tesla V100 with 32G GPU memory and found the highest resolution in which we can train our model under the limitation of GPU memory.

\begin{figure}[tb]
    \centering
    \includegraphics[width=\linewidth]{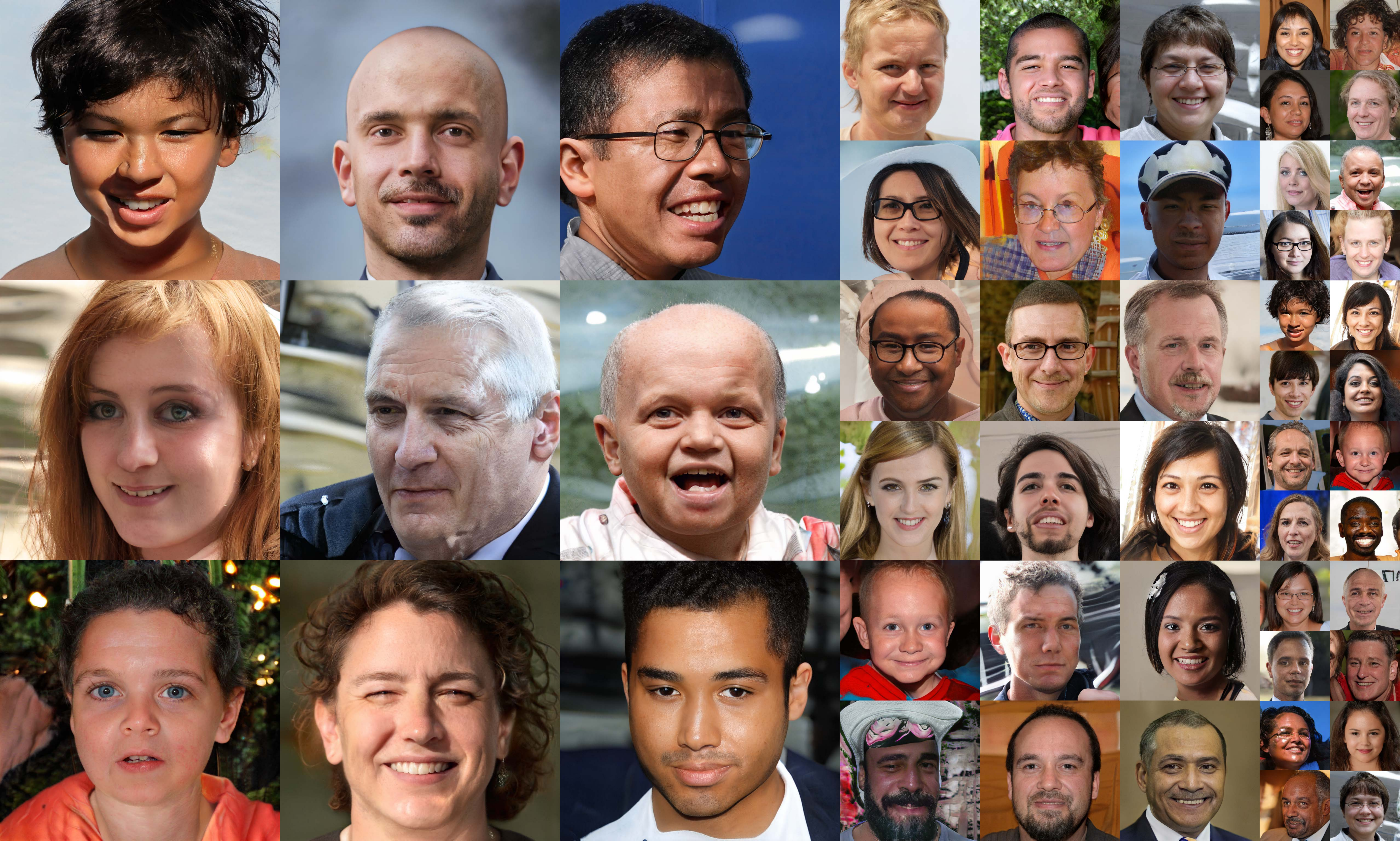}
    \caption{Sampled images from MSStyleGAN with $1024^2$, $512^2$ and $256^2$ resolutions. The larger resolution is shown at larger size.}
    \label{fig:appx-ms-1024}
\end{figure}

\begin{figure}[tb]
    \setlength{\abovecaptionskip}{3pt}
    \centering
    \small
    \includegraphics[width=0.96\linewidth]{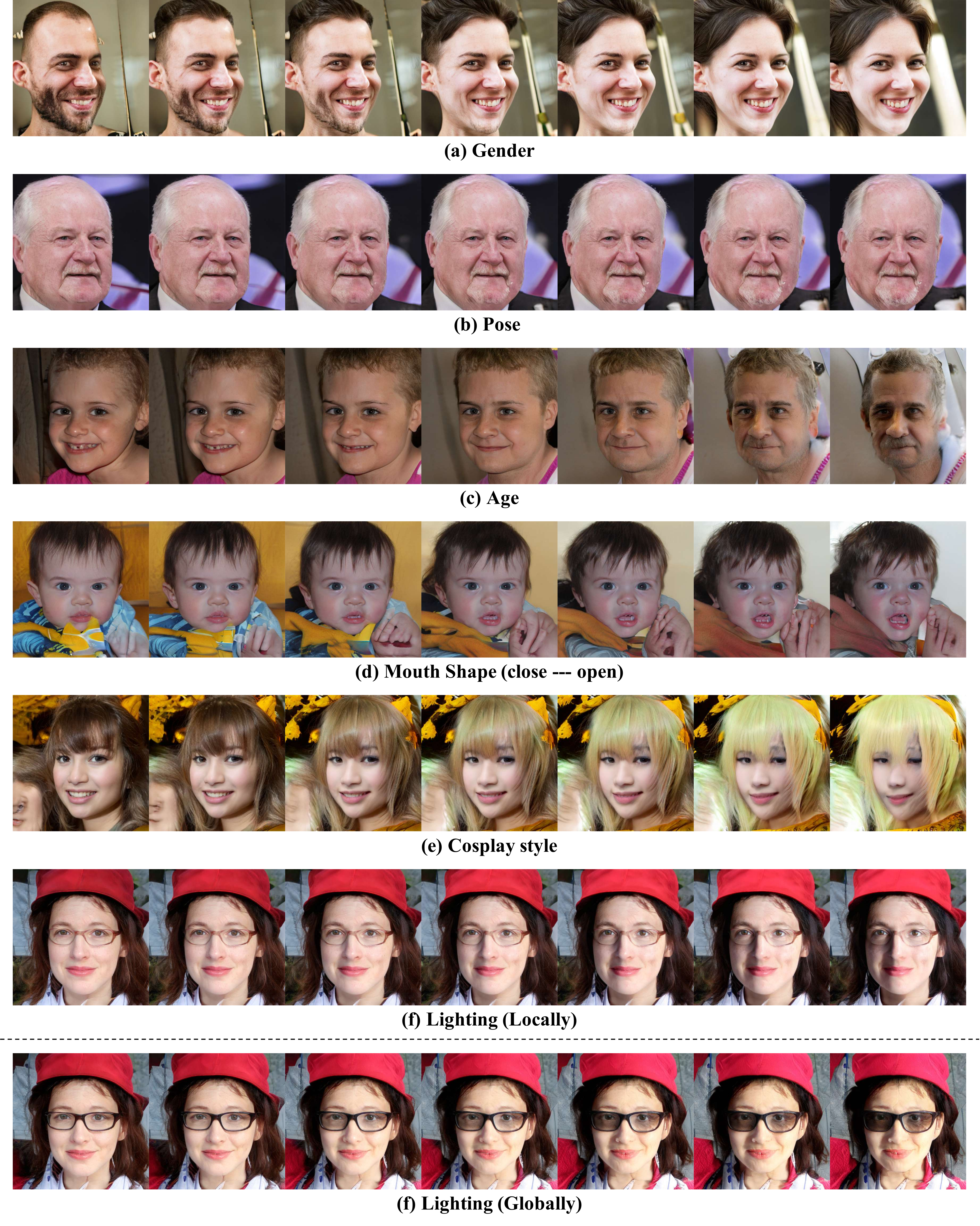}
    \caption{Image manipulation in $512 \times 512$ resolution with a $256^2$ StyleGAN2 trained in MS-PIE.}
    \label{fig:appx-manipulate}
\end{figure}

For $256^2$ StyleGAN2 with a channel multiplier of two, $896\times 896$ is the highest resolution in which we can achieve a competitive FID of 4.10. We adopt three different scales of $256^2$, $512^2$, and $896^2$ and the synthesized images are shown in Fig~\ref{fig:appx-ms-896}.
Reducing the channel multiplier to one, we can train the lite StyleGAN2 model with $256^2$, $512^2$, and $1024^2$ scales in MS-PIE.
Even if only containing six lite convolutional blocks, as shown in Fig.~\ref{fig:appx-ms-1024}, the lite generator can achieve compelling generation quality with an FID of 6.24.

\subsection{Image Manipulation}
\label{sec:appx-manipulation}

To further verify the effectiveness of our MS-PIE in image editing, we customize a manipulation algorithm for $512\times 512$ image manipulation with a single $256^2$ StyleGAN2.
Based on the best training configuration in Tab. 5(f) in Sec.~\ref{sec:exp-ms}, we will demonstrate its effectiveness.
Achieving image manipulation in higher resolution with the $256^2$ backbone is not trivial.
We have tried the closed-form factorization method~\cite{shen2020closed} in our generators but found that it cannot manage to control the image style. 
After analyzing the results and improving the closed-form factorization method, a more convenient and effective manipulation algorithm is proposed for the generators trained in MS-PIE.

Firstly, unlike the original method, we directly concatenate all of the weights in the style encoding layer from the upsampling convolutional blocks and compute the eigenvector for the large weight matrix.
In our experiments, we found that such eigenvectors from the large weight matrix can control the attributes of the output.
Furthermore, different eigenvectors tend to control different attributes separately.
As shown in Fig.~\ref{fig:appx-manipulate}, the third eigenvector only controls the gender attribute while the $10$-th eigenvector only controls the pose attribute.

The eigenvectors are computed from the large weight matrix containing weights from all of the convolutional blocks.
Such eigenvectors can be applied to each convolutional block or only several blocks.
For example, we can \bfit{globally} adopt the third eigenvector in each convolutional block. However, as shown in Fig.~\ref{fig:appx-manipulate}, this eigenvector will not have any influence on other attributes, \eg, pose, and mouth shape.
The $15$-th eigenvector controls the lighting of the output and the $8$-th and the $9$-th convolutional block are much sensitive to this eigenvector.
As shown in Fig.~\ref{fig:appx-manipulate}, the lighting attribute can be well controlled by \bfit{locally} applying the $15$-th eigenvector in the $8$-th and $9$-th convolutional block. 
Once \bfit{globally} applying the $15$-th eigenvector in each convolutional block, we observe that other attributes, like the color of the glass, will be influenced by the lighting.

\section{SinGAN}
\label{appx:singan}

\subsection{Implementation Details}

In SinGAN, we strictly follow the original training configuration to study the padding effects and the impacts of spatial inductive bias.
For all of the training samples, we adopt the minimum size of 25 at the start stage while the final size of the final stage is set according to the original scale of the training image. 
As for padding removal, we carefully discard the padding from the input noise feature and the input images at each stage.
In the experiments on the Cartesian grid, we adopt the Cartesian grid with two channels as the input positional encoding.
Without other specifications, the number of total channels in sinusoidal positional encoding is set to 8. 
We use 16 channels in the sample of `\textit{Bohemian Rhapsody}'.

\begin{figure}
    \centering
    \includegraphics[width=\linewidth]{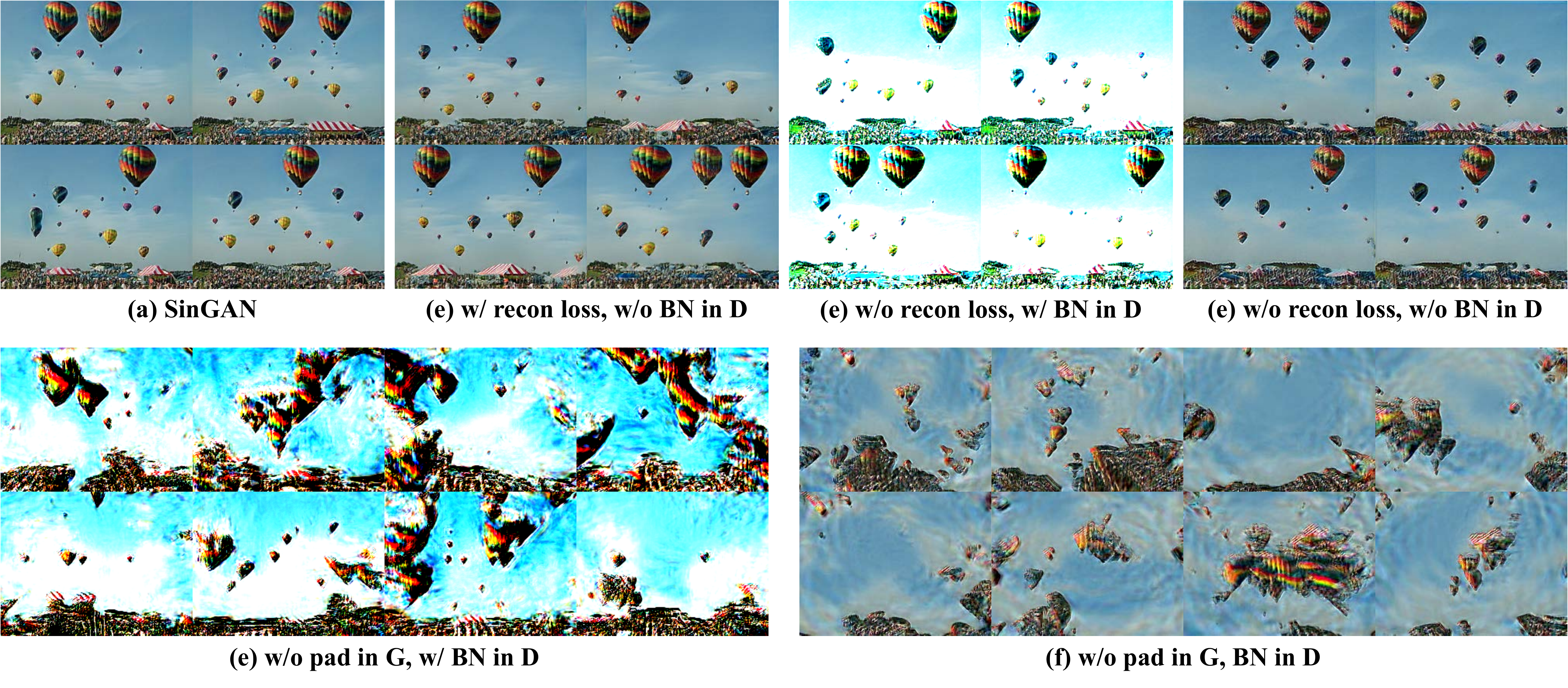}
    \caption{The effects of adopting batch normalization in discriminator and the reconstruction loss on SinGAN.}
    \label{fig:appx-bn-singan}
\end{figure}

In the experiments with padding-free SinGAN, we modify the architecture of the discriminator by removing the batch normalization layer. 
This is because the channel-wise normalization layer destroys the learned relationship between the different channels, which brings a significant color shift in the output image.
%
%
In Fig.~\ref{fig:appx-bn-singan}, we presents an ablation study about the color shift phenomenon.
Firstly, as shown in Fig.~\ref{fig:appx-bn-singan}(b), removing the batch normalization layer in the discriminator will not influence the generated spatial structure, despite marginal negative effects on the texture quality.
Once discarding the reconstruction loss in SinGAN, the generator performs a significant color shift with unreasonable brightness in the results in Fig.~\ref{fig:appx-bn-singan}(c).
Meanwhile, we observe that the spatial structure can be retained without any reconstruction supervision.
Furthermore, based on the model without reconstruction loss, we remove the batch normalization layer in the discriminator.
Surprisingly, the results in Fig.~\ref{fig:appx-bn-singan}(d) recover the original brightness, indicating that the reconstruction loss indeed plays a role in correcting the color shift brought by adopting batch normalization in the discriminator.
Finally, in Fig.~\ref{fig:appx-bn-singan}(e) and Fig.~\ref{fig:appx-bn-singan}(f), we show the results from padding-free SinGAN trained with different discriminators, \ie, with batch normalization and without batch normalization.
With a padding-free generator, adopting batch normalization in the discriminator only brings the color shift in results while the spatial structure cannot be captured. 
Thus, removing batch normalization in the discriminator does not influence the generated spatial structure that we care about in this work.

\begin{figure}[tb]
    \centering
    \includegraphics[width=\linewidth]{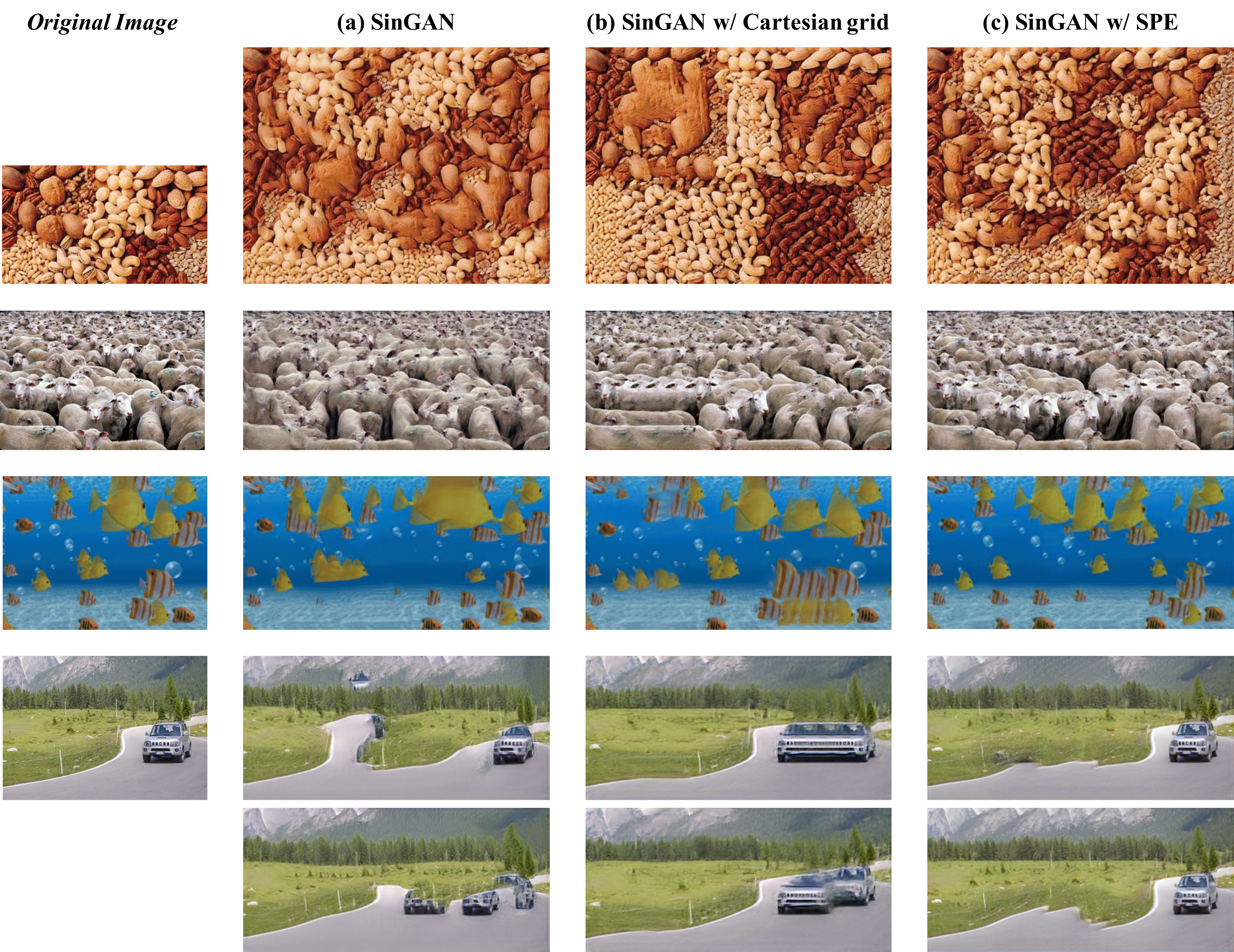}
    \caption{More results for SinGAN with different positional encodings. }
    \label{fig:appx-singan-exp}
\end{figure}

\subsection{More Qualitative Results from SinGAN}
Figure \ref{fig:appx-singan-exp} presents more comparison for the effects of adopting different spatial inductive bias on SinGAN.
The first case with different nuts verifies that the Cartesian grid can naturally keep the global structure retained, \eg, the regular arrangements as the same as the original image.
However, such spatial inductive bias is not suitable to perform multi-scale synthesis in the last case with a car.
It is better to choose the sinusoidal positional encoding for a faithfully detailed structure and realistic patch recurrence.
On the contrary, without a clear or balanced spatial bias over the whole image space, the standard SinGAN cannot manage to perform high-quality internal sampling.

\end{document}